\documentclass[5p,twocolumn]{elsarticle}

\usepackage{amssymb}
\usepackage{latexsym}

\usepackage{url}
\usepackage{xcolor}
\usepackage{hyperref}
\usepackage{amsmath}
\usepackage{booktabs}
\usepackage{ulem}
\usepackage{color}
\definecolor{newcolor}{rgb}{.8,.349,.1}

\journal{Medical Image Analysis}

\begin{document}

\begin{frontmatter}

\title{MIDeepSeg: Minimally Interactive Segmentation of Unseen Objects from Medical Images Using Deep Learning}
\author[1]{Xiangde Luo}

\author[1]{Guotai~Wang\corref{mycorrespondingauthor}}
\cortext[mycorrespondingauthor]{Corresponding author}
\ead{guotai.wang@uestc.edu.cn}

\author[2]{Tao Song}

\author[3]{Jingyang Zhang}
\author[5]{Michael Aertsen}

\author[5,6,7]{Jan~Deprest}

\author[4]{Sebastien~Ourselin}

\author[4]{Tom~Vercauteren}

\author[1,2]{Shaoting~Zhang}

\address[1]{School of Mechanical and Electrical Engineering, University of Electronic Science and Technology of China, Chengdu, China}

\address[2]{SenseTime Research, Shanghai, China}

\address[3]{School of Biomedical Engineering, Shanghai
Jiao Tong University, Shanghai, China}

\address[4]{School of Biomedical Engineering and Imaging Sciences, King's College London, London, UK}

\address[5]{Department of Radiology, University Hospitals Leuven, Leuven, Belgium
}

\address[6]{Department of Obstetrics and Gynaecology, University Hospitals Leuven, Leuven, Belgium}

\address[7]{Institute for Women’s Health, University College London, London, UK}

\begin{abstract}
Segmentation of organs or lesions from medical images plays an essential role in many clinical applications such as diagnosis and treatment planning. Though Convolutional Neural Networks (CNN) have achieved the state-of-the-art performance for automatic segmentation, they are often limited by the lack of clinically acceptable accuracy and robustness in complex cases. Therefore, interactive segmentation is a practical alternative to these methods. However, traditional interactive segmentation methods require a large amount of user interactions, and recently proposed CNN-based interactive segmentation methods are limited by poor performance on previously unseen objects. To solve these problems, we propose a novel deep learning-based interactive segmentation method that not only has high efficiency due to only requiring clicks as user inputs but also generalizes well to a range of previously unseen objects. Specifically, we first encode user-provided interior margin points via our proposed exponentialized geodesic distance that enables a CNN to achieve a good initial segmentation result of both previously seen and unseen objects, then we use a novel information fusion method that combines the initial segmentation with only few additional user clicks to efficiently obtain a refined segmentation. We validated our proposed framework through extensive experiments on 2D and 3D medical image segmentation tasks with a wide range of previous unseen objects that were not present in the training set. Experimental results showed that our proposed framework 1) achieves accurate results with fewer user interactions and less time compared with state-of-the-art interactive frameworks and 2) generalizes well to previously unseen objects.
\end{abstract}

\begin{keyword}
Interactive image segmentation\sep convolutional neural network\sep geodesic distance \sep generalization \end{keyword}
\end{frontmatter}

\section{Introduction}Accurate and robust segmentation of organs or lesions from medical images plays an essential role in many clinical applications such as diagnosis and treatment planning~\citep{Zhao2013, masood2015survey}. Although automatic segmentation methods have been studied for many years, it remains challenging for them to obtain a consistently precise segmentation in cases with large anatomical variation and complex pathologies~\citep{Wang18}. This is mainly due to the inherent limitations of medical images, such as low contrast, different imaging and segmentation protocols, and variations among patients~\citep{Wang18}. In contrast, interactive segmentation methods, which leverage the user's knowledge and experience to obtain a more accurate and robust result, are more practical and widely used in clinical applications~\citep{Zhao2013, masood2015survey, Wang18}. 

\par A desirable interactive segmentation tool should 1) achieve accurate segmentation results with as few user inputs as possible, leading to reduced burdens on the user; 2) have high efficiency so that the user can get real-time response, even when dealing with volumetric data; 3) generalize well to different objects so that it is ready-to-use for new objects or image modalities. However, existing interactive segmentation methods rarely satisfy all these often competing requirements. Many traditional interactive methods use low-level features (e.g., gray level or color distribution) for image segmentation~\citep{Hu2019}, such as Graph Cuts~\citep{Boykov2002}, ITK-SNAP~\citep{Yushkevich2006}, GeoS~\citep{Criminisi2008}, Random Walks~\citep{Grady2006} and GrowCut~\citep{Vezhnevets2005}. As low-level features cannot effectively distinguish the object from the background in many situations with low contrast~\citep{Hu2019}, these methods often require a large amount of user interactions and long user time to obtain reliable results. To reduce the amount of annotations required from the user to build an adequate foreground/background model, machine learning has been widely used to perform interactive segmentation. For example, SlicSeg~\citep{Wang2016} and DyBaORF~\citep{Guotai2016} use an Online Random Forest (ORF) to segment the placenta from Magnetic Resonance Imaging (MRI) volume. GrabCut~\citep{Rother2004} uses Gaussian Mixture Models (GMMs) to estimate the foreground and background distributions. It obtains an initial result by a user-provided bounding box around the region of interest and allows additional interactions for refinement. \cite{Wang2014} used active learning to actively select candidate regions for querying the user to obtain much informative user feedback and thus reduced user interactions. These algorithms perform better than traditional methods without machine learning, but they are limited by the use of hand-crafted features~\citep{Wang2018, Wang18}. As a result, they still require a considerable amount of user interactions for accurate segmentation. 

\par Recently, with the ability to learn high-level semantic features automatically, deep learning with Convolutional Neural Networks (CNNs) has achieved state-of-the-art performance for image segmentation~\citep{Shen2017, Litjens2017}. To take advantage of the good representation ability of CNNs and overcome the limited accuracy and robustness of the automatic CNNs, some deep learning-based interactive segmentation tools~\citep{Wang18, Hu2019, Wang2018, Rueckert2016, Xu2016, Maninis2017, sakinis2019interactive} have recently been proposed. The methods of~\citet{Hu2019}, \citet{Xu2016} and \citet{Maninis2017} are designed to segment 2D RGB images interactively and lack evaluation on medical images with low contrast and ambiguous boundaries. \citet{castrejon2017annotating} and \citet{acuna2018efficient} integrated reinforcement learning and graph neural networks into a unified polygon-based interactive segmentation framework, where the user is allowed to drag a point on the polygon for refinement, but its ability to deal with objects with complex shapes and 3D medical images is limited. 

In contrast, DeepIGeoS~\citep{Wang18}, IFSeg~\citep{sakinis2019interactive}, DeepCut~\citep{Rueckert2016} and BIFSeg~\citep{Wang2018} are specially designed to segment medical images.
DeepCut~\citep{Rueckert2016} uses a set of user-provided bounding box as sparse annotations to train CNNs for the segmentation of fetal brain and lung from fetal MRI.~\citet{roth2019weakly,roth2020going} combined extreme points~\citep{Maninis2017} with random walkers~\citep{Grady2006} for weakly supervised 3D medical image segmentation. Even though this method and DeepCut~\citep{Rueckert2016} reduced the annotation cost significantly, it was designed for weakly supervised model training over a large dataset rather than interactively editing a single segmentation result at test time.  \citet{raju2020user} further used extreme points in a user-guided domain adaptation method for pathological liver segmentation.
DeepIGeoS~\citep{Wang18} performs user-friendly interactive segmentation by combining CNNs and user-provided scribbles, where a CNN is used to obtain an initial segmentation and another CNN accepts additional user interactions for refinement. However, DeepIGeoS can only deal with objects present in the training set and lacks of adaptability to previously unseen objects. Following~\citet{Xu2016}, IFSeg~\citep{sakinis2019interactive} takes user clicks and the raw image as input for interactive medical image segmentation. Despite the fact that the framework is easy to use, it generalizability was only validated with a single previously unseen structure, and the ability to deal with various unseen objects in different modalities was not shown.
BIFSeg~\citep{Wang2018} exploits user-provided bounding boxes and image-specific fine-tuning to segment some unseen objects, but it is limited by dealing with only few unseen objects in the same image modality or similar context and requiring time-consuming fine-tuning for each test image.
Therefore, novel interactive frameworks  for medical image segmentation with higher efficiency and generalizability is highly desirable. 

\par Besides, a practical problem for CNN-based interactive segmentation methods is to effectively encode user interactions, as different encoding strategies have a large impact on the interactive segmentation performance. Most of existing works encode user interactions by transforming them into a cue map, such as Euclidean distance map~\citep{Hu2019, Xu2016,Li2018, Benenson2019, Hao2019}, Gaussian heatmap~\citep{Maninis2017, Cvpr2019}, and iso-contours derived from user clicks~\citep{Khan2019}. However, these encoding methods do not take the image context information into account. In contrast, the geodesic distance transform is spatially smooth and contrast-sensitive to encode user interactions~\citep{Criminisi2008, Bai2009, Price2010}. DeepIGeoS~\citep{Wang18} uses geodesic distance transform with a specially designed threshold to deal with user-provided interactions. However, it is time-consuming to find an appropriate threshold value to truncate the generated geodesic distance map when dealing with different objects. We assume that a context-aware and parameter-free encoding method is helpful for improving the segmentation accuracy and generalizability.  
\begin{figure*}
	\centering
	\includegraphics[width=1.0\linewidth]{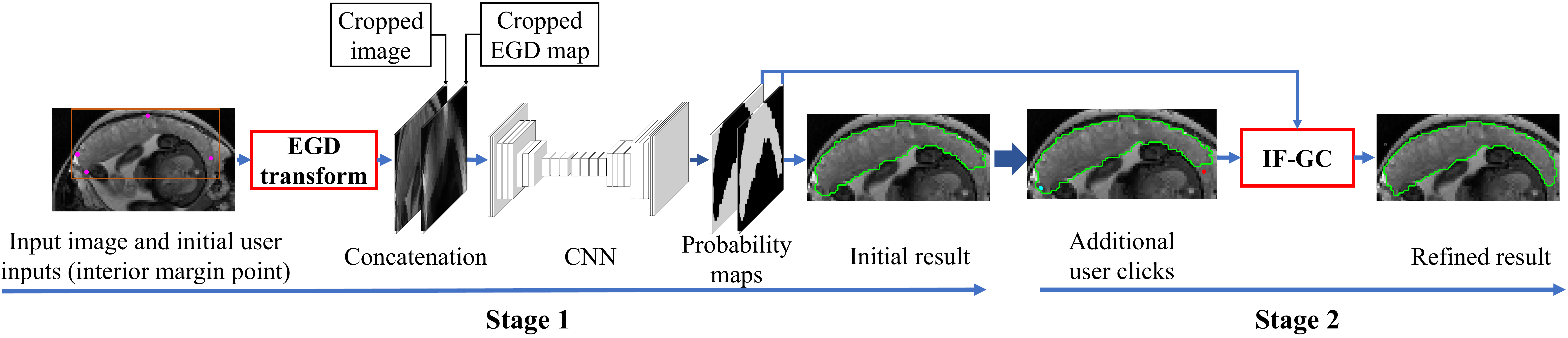}
	\caption{Pipeline of the proposed Minimally Interactive Deep learning-based Segmentation framework (MIDeepSeg). Stage 1: User-provided interior margin points are encoded by Exponentialized Geodesic Distance (EGD) maps to guide a CNN to obtain an initial result. Stage 2: Refining the initial segmentation based on additional user clicks and our proposed Information Fusion followed by Graph Cut (IF-GC). Note that this framework is ready to use for segmentation of previously unseen objects without the need of extra fine-tuning or re-training.
	}
	\label{fig:flowchart}
\end{figure*}
\par To tackle the above-mentioned challenges, we propose a new generalizable framework for more intelligent and accurate interactive segmentation of 2D and 3D medical images, which aims at not only obtaining high performance and efficiency for segmentation of previously seen objects, but also achieving high generalization to a range of previously unseen objects. Our method takes advantages of CNNs and only requires few clicks as user interactions. We present a new way to encode user interactions based on Exponentialized Geodesic Distance (EGD) transform, which is context-aware and parameter-free and helps to improve the segmentation obtained by the CNN. We also propose an information fusion method that efficiently fuses additional user clicks with the initial segmentation to obtain a refined segmentation. Differently from existing interactive medical image segmentation frameworks~\citep{Wang2018, Wang18, Rueckert2016}, our method is more efficient as it only works on a sub-region of the image and does not need to train an additional CNN on the fly for the refinement. Moreover, we validate the effectiveness of this framework with a large range of previously seen and unseen objects. The superiority of our method over existing interactive segmentation methods is validated with five types of 2D unseen objects and four types of 3D unseen objects from different types of image contexts and modalities.

\begin{figure}
	\centering
	\includegraphics[width=1.0\linewidth]{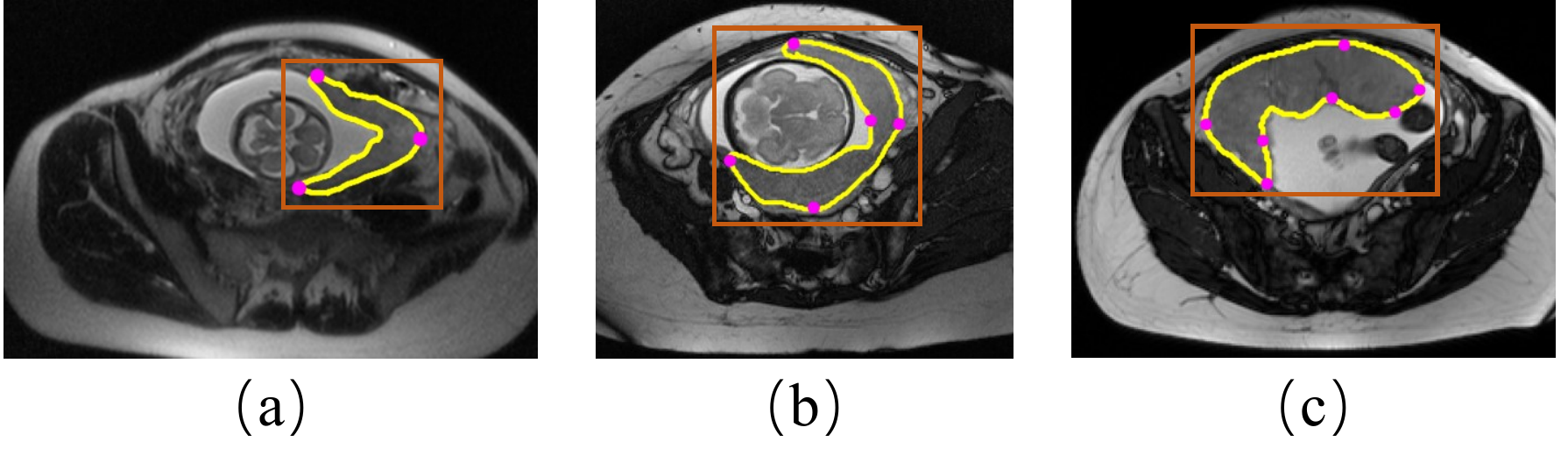}
	\caption{Simulation of interior margin points on training images for different shapes of placenta. Fuchsia: simulated clicks on placenta edge. Brown: interior margin points-derived relaxed bounding box. Yellow: ground truth.}
	\label{fig:simulated-points}
\end{figure}
\section{Methods}The proposed Minimally Interactive Deep learning-based Segmentation framework is referred to as MIDeepSeg and illustrated in Fig.~\ref{fig:flowchart}. It consists of two stages. In the first stage, the user provides few clicks near the boundary (i.e., interior margin point) of the target object. These points are used to infer a relaxed bounding box to crop the input image. Based on the cropped image, all user-provided interior margin points are converted to a cue map based on our proposed EGD transform. Then, the cue map is concatenated with the cropped input image as the input of a CNN to obtain an initial segmentation result. In the second stage, the user provides some additional clicks to indicate mis-segmented regions, and a refined result is obtained by our proposed Information Fusion followed by Graph Cuts (IF-GC). At test time, the refinement step can run several times until the result is accepted by the user. After training with a small set of objects, our framework is ready to use for the segmentation of previously unseen objects without the need for fine-tuning or re-training that is time-consuming and requires additional annotations.

\subsection{User interaction based on interior margin points}\label{sec2.1:user-interactions}
Many existing CNN-based interactive segmentation frameworks use scribbles~\citep{Wang2018} or bounding boxes~\citep{Rueckert2016} or both~\citep{Wang18} as interactive cues. They need the user to drag the cursor carefully, which requires a lot of user's efforts~\citep{Papadopoulos2017, Maninis2017}. Using clicks as user interactions is a more user-friendly and effective way as demonstrated by previous works~\citep{Xu2016, Maninis2017, Cvpr2019, Papadopoulos2017}. Recently,~\citet{Maninis2017} proposed a framework that only needs the user to provide clicks for extreme points (i.e, left-, right-, top- and bottom-most pixels) of an object for RGB image segmentation, which reduces the amount of user interactions substantially. However, in medical images, the accurate extreme points are hard and time-consuming to find, which increases the burden on the user, since the target organs or lesions have a large variability in the size and shape across different patients or imaging protocols, especially in 3D volumetric data. In addition, for irregular and concave shapes, extreme points are not enough to capture the main shape of the object (as shown in Fig.~\ref{fig:simulated-points}), which can limit the performance of the CNN. 

\begin{figure*}
	\centering
	\includegraphics[width=1.0\textwidth]{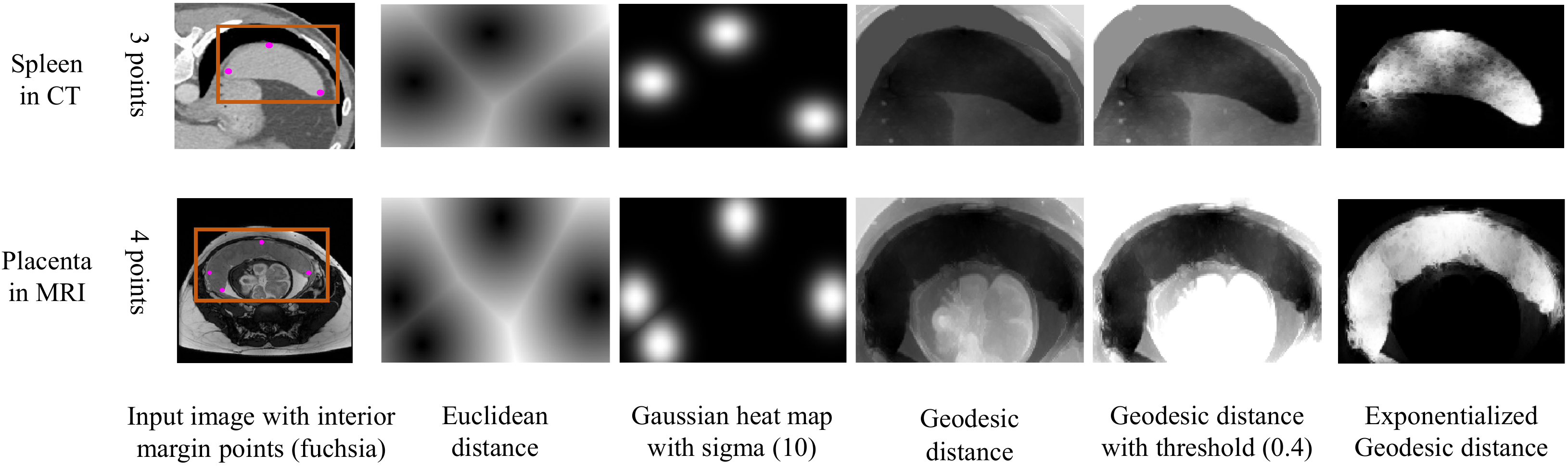}
	\caption{Visual comparison of different cue maps generated from user-provided interior margin points. (Fuchsia: interior margin points. Brown: inferred relaxed bounding box)}
	\label{fig:distance-tansforms}
\end{figure*}
To alleviate these limitations, we  propose to use interior margin points as user interactions, where the user only needs to provide some clicks that are in the inner side and close the boundary of the target. Compared with DEXTR~\citep{Maninis2017} that uses at most four extreme points and optionally with one extra point, our interior margin points can provide more shape information for different types of organs with complex and irregular shapes. In addition, putting clicks exactly on the object boundary and even extreme points is hard for users at test time, and relaxing the clicks to inner side of the boundary makes the interactions more friendlier and convenient to implement, which tolerates inaccurate clicks. We relax these points towards the inside region because an exponentialized geodesic distance transform of these interior margin points can be a good approximation of the saliency map of the target object, as shown in  Fig.~\ref{fig:distance-tansforms}. Therefore, interior margin points bring potential advantages in guiding CNNs to deal with different types of unseen objects as well.

\par During training, all interior margin points for each object were automatically simulated based on the ground truth mask and edge detector~\citep{Harris1988}. The interior margin points are generated based on two rules: First, these points should be located in the object and near the boundary. Second, a relaxed bounding box determined by these points should cover the entire object region. Therefore, we simulate user interactions for a training image in two steps. 1) To ensure that the relaxed bounding box covers the whole region of interest, few points on the ground truth boundary (three or four for 2D objects, five or six for 3D objects) close to the extreme points~\citep{Maninis2017} of the target object are selected. Then, we randomly sample $n$ points from remaining boundary points of the target to provide more shape information, where $n$ is a random number from 0 to 5. 2) To simulate real user clicks that may not be accurately positioned on the object boundary, all these points obtained in step 1 are slightly moved towards the inner side of the boundary by several pixels/voxels to obtain our interior margin points. We moved simulated points towards the inner side of the target object as the users are asked to put the interior margin points in the inner side of the boundary as well. Then the bounding box determined by these points is relaxed by several pixels/voxels to include some background region. Examples of simulated 2D interior margin points and relaxed bounding boxes on training images are shown in Fig.~\ref{fig:simulated-points}. In test stage, the user is required to provide the interior margin points in such a way that they satisfy the above two rules. The relaxed bounding box determined by user interactions is expanded with a small margin to include some contextual information.

\subsection{Exponentialized geodesic distance transform}It is critical for CNN-based interactive methods to encode user interactions efficiently. A desirable encoding method should take image context into account and can be combined with CNNs directly without any manually designed parameters. However, existing interaction encoding methods such as Euclidean distance transform~\citep{Xu2016, Li2018}, Gaussian heatmap~\citep{Maninis2017, Cvpr2019}, iso-contours~\citep{Khan2019} and geodesic distance transform~\citep{Wang18} do not have these merits at the same time. To deal with this problem, we propose a context-aware and parameter-free encoding method: Exponentialized Geodesic Distance (EGD) transform, which is a combination of geodesic distance transform and exponential transform. 
\begin{figure*}
	\centering
	\includegraphics[width=0.7\linewidth]{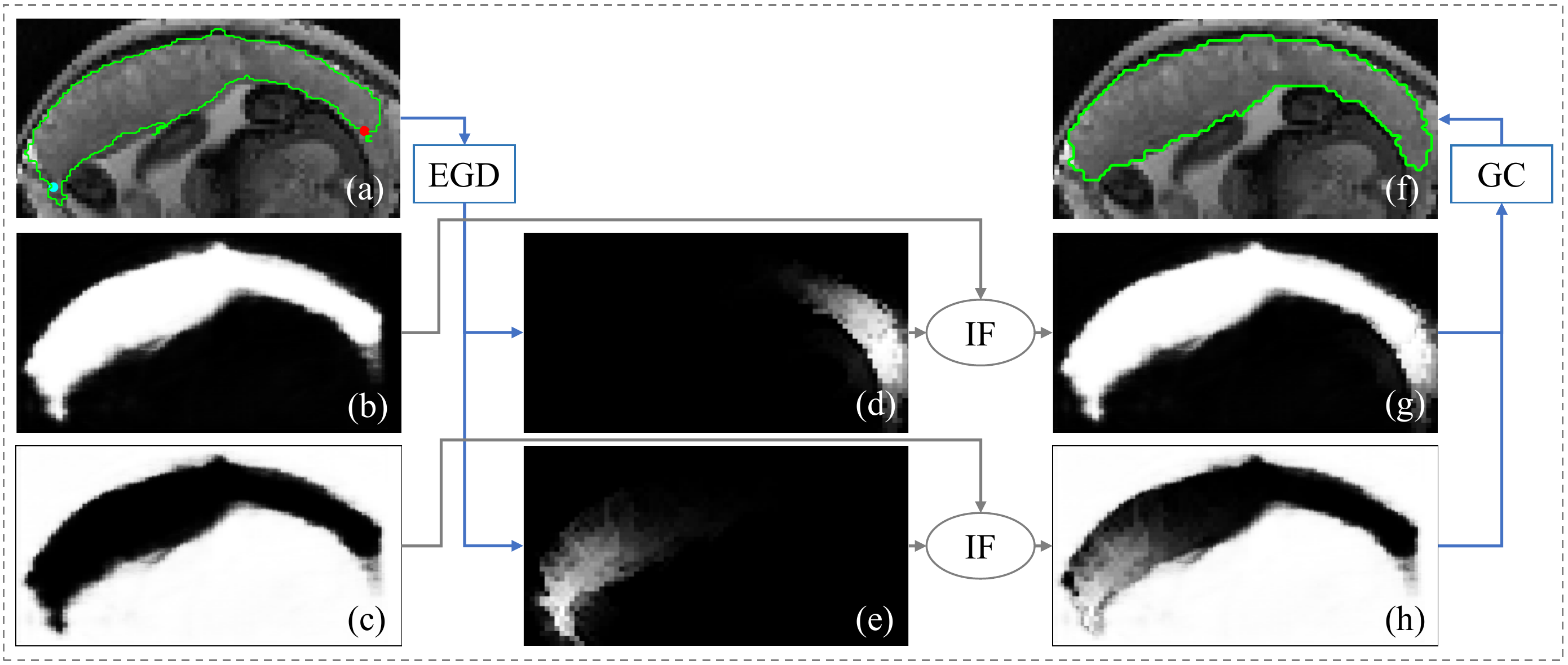}
	\caption{Illustration of refinement by information fusion. (a) The user provides clicks to indicate under-segmentation(red) and over-segmentation(cyan) regions. (b) and (c) are initial segmentation foreground and background probability maps obtained by CNN in the first stage, respectively. (d) and (e) are cue maps based on foreground and background refinement clicks and EGD transformation, respectively. (g) and (h) are calibrated foreground and background probability maps, respectively. (f) is refined segmentation result. (IF: Information Fusion; EGD: Exponentialized geodesic distance transformation; GC: Graph Cut)}
	\label{fig:refined_method}
\end{figure*}

Suppose $\mathcal{S}_s$ represent the set of pixels/voxels belonging to the simulated interior margin points in the training stage or user-provided interior margin points in the testing stage. Let ${i}$ be a pixel/voxel in the input image $I$, then the unsigned EGD from ${i}$ to $\mathcal{S}_s$ is:
\begin{eqnarray}
  {EGD}(i, S_s, I) = \min \limits_{ j \in S_s} e^{-\mathcal{D}_{geo}({i},{j},{I})}
\end{eqnarray}
\begin{eqnarray}
\mathcal{D}_{geo}(i, j, I) = \min \limits_{p \in \mathcal{P}_{i,j}} \int_0^1 \left \| \nabla I(p(n)) \cdot v(n) \right \| dn
\end{eqnarray}where $\mathcal{P}_{i,j}$ is the set of all paths between pixels/voxels $i$ and $j$. $p$ is one feasible path and it is parameterized by ${n} \in \ \left [0, 1  \right ]$. ${v}(n) = {p}'({n}) / \left \| {p}'({n})\right \|$ is a unit vector that is tangent to the direction of the path. Note that the EGD here is defined for scalar images but can easily be extended to vector-valued (i.e., multi-channel or multi-modal) images. Fig.~\ref{fig:distance-tansforms} shows some examples of cue maps obtained by different encoding methods applied to some interior margin points. It can be observed that EGD-based cue map differentiates the foreground from the background better than those based on the other encoding methods. Therefore, it has the potential to provide more shape, position and context information to guide the CNN to obtain a good initial segmentation result.

\subsection{Initial segmentation based on cue map and CNN}
In this paper, we focus on designing an efficient and general framework to deal with seen and unseen objects from different types of images. Therefore, our framework does not rely on a specific design of CNN structure. To demonstrate its utility, we use adapted 2D-UNet~\citep{Ronneberger2015} and 3D-UNet~\citep{Cicek2016} for 2D and 3D segmentation, respectively. We replace the batch normalization layers with instance normalization layers that has a better adaptability to different kinds of images, and reduce the feature channel numbers by four times to balance the performance, memory cost, and time consumption. In the training stage, all interior margin points and relaxed bounding boxs are automatically simulated based on the ground truth label, as described in Section~\ref{sec2.1:user-interactions}. Then all interior margin points are converted into a cue map that is concatenated with the cropped input image as the input of the CNN, as shown in Fig.~\ref{fig:simulated-points}. In the testing stage, the user is asked to provide interior margin points for a given target. Then, the CNN can give an initial segmentation result. To correct the mis-segmentation, we use a refinement stage with information fusion between the initial segmentation and additional user clicks, as described in the following.
\subsection{Refinement based on information fusion between initial segmentation and additional user clicks}For deep learning-based interactive segmentation, it is important to support refinement of an initial segmentation. Existing methods either require an additional model for refinement~\citep{Wang18, castrejon2017annotating, acuna2018efficient, zhou2019interactive, Liao_2020_CVPR} or need to fine-tune the pre-trained model for a specific image~\citep{Wang2018}. However, these refinement methods are time and memory consuming, and not ready-to-use for unseen objects. In addition, ~\citet{chen2017deeplab} and~\citet{kamnitsas2016deepmedic} used CRF~\citep{Lafferty2001} to refine CNN's prediction automatically. However, these CRF-based~\citep{Lafferty2001} refinement methods~\citep{chen2017deeplab,kamnitsas2016deepmedic} were not designed for interactive segmentation. Differently from these methods, we propose an efficient and simple refinement method based on a novel method for information fusion between initial segmentation and additional user interactions, which generalizes better to previously unseen objects without extra fine-tuning and re-training. Fig.~\ref{fig:refined_method} shows an illustration of our information fusion method. 
\par In the refinement stage, the user is asked to provide some additional clicks to indicate mis-segmented foreground and background regions, respectively. To efficiently encode these new interactions, we use the proposed EGD transform again to get two additional interaction-derived cue maps: $E^f$ and $E^b$ are cue maps based on EGD of user-provided foreground and background clicks for refinement, respectively. Note that, we do not reuse the initial EGD map obtained in the first stage directly, but  combine the initial interior margin points with refinement clicks for calculating the new EGD maps in the refinement step. The values of $E^f$ and $E^b$ are in the range of $ \left [0, 1  \right ]$ and represent the similarity between each pixel and foreground/background clicks. Let $P^f$ and $P^b$ denote the initial foreground and background probability map obtained by the CNN, and $i$ represent a pixel/voxel in the input image $I$. The information fusion strategy is proposed to refine the $P^f$ and $P^b$ according $E^f$ and $E^b$. Specifically, we aim to automatically emphasis on $E^f$ and $E^b$ when pixel $i$ is close to  the refinement clicks, otherwise $P^f$ and $P^b$ tend to keep unchanged. We define user-calibrated foreground ($R^f_i$) and background ($R^b_i$) probability for pixel $i$ as:
\begin{eqnarray}
    {E}^{f}_{i} = \frac{e^{-D^{f}_{i}}}{e^{-D^{f}_{i}} + e^{-D^{b}_{i}}}
\end{eqnarray}

\begin{eqnarray}
    {E}^{b}_{i} = \frac{e^{-D^b_i}}{e^{-D^f_i} + e^{-D^b_i}}
\end{eqnarray}

\begin{eqnarray}\label{eq:eq6}
    {R}^{f}_{i} = (1 - \alpha_i) * P^f_i + \alpha_i * E^f_i
\end{eqnarray}

\begin{eqnarray}
    {R}^{b}_i = (1 - \alpha_i) * P^b_i + \alpha_i * E^b_i
\end{eqnarray}

\begin{eqnarray}
    \alpha_i = e^{-\min(D^f_i, D^b_i)}
\end{eqnarray}
where $\alpha_i \in [0,1]$ is an automatic and adaptive weighting factor. When $i$ is close to the clicks, $\alpha_i$ is close to 1.0, and $R^f_i$ ($R^b_i$) is more affected by $E^f_i$ ($E^b_i$). If no clicks are provided for the foreground (background), we set the corresponding $D^f_i$ or $D^b_i$ to a constant value. Let $C^f$ and $C^b$ denote the clicks for foreground and background, respectively, so the entire set of clicks is $C = {C}^{f}$ $\cup$ ${C}^{b}$. Let ${c}_{i}$ denote the user-provided label of a pixel in the clicks, then we have ${c}_{i}$ = 1 if $i$~$\in$~${C}^{f}$ and ${c}_{i}$ = 0 if $i$~$\in$~${C}^{b}$. We integrate ${R}^{f}$ and ${R}^{b}$  into a Conditional Random Field (CRF) to get the refined segmentation:

\begin{eqnarray}\label{eq:eq7}
&{E} = \sum \limits_{i} {\phi}({y}_{i}|I)+\lambda \cdot \sum \limits_{i, j} {\psi}({y}_{i}, {y}_{j}|I)\\ & \text{subject to}: {y}_{i} = {c}_{i} \; \text{if} \; {i} \; \in \; {C}
\end{eqnarray}where $\phi$ and $\psi$ are the unary and pairwise energy terms, respectively. $\lambda$ specifies a relative weight between $\phi$ and $\psi$. In this paper:
\begin{eqnarray}\label{eq:eq8}
   {\phi}({y}_{i}|I) = -({y}_{i}\log({r}_{i}) + (1 - {y}_{i})\log(1 - {r}_{i}))
\end{eqnarray}

\begin{eqnarray}\label{eq:eq9}
    {\psi}({y}_{i}, {y}_{j}|I) \propto \exp( - \frac{({I}_{i} - {I}_{j})^{2}}{2\sigma^{2}}) \cdot \frac{1}{{dist}_{ij}}
\end{eqnarray}where $r_{i}$ denotes the value of pixel $i$ in ${R}^{f}$, and ${y}_{i} = 1$ if ${i}$ belongs to the foreground and 0 otherwise. $I_i$ and $I_j$ mean the intensity of pixel $i$ and $j$ in image image $I$, respectively. ${dist}_{ij}$ is the Euclidean distance between pixel/voxel $i$ and $j$. $\sigma$ is a parameter to control the effect of intensity difference. In this paper, the CRF~\citep{Lafferty2001} problem in Eq.~\eqref{eq:eq7} is submodular and can be solved by Graph Cut through max-flow/min-cut~\citep{Boykov2002}.

\subsection{Implementation details}
We implemented the U-Net and 3D U-Net for 2D and 3D images segmentation by Pytorch~\citep{pytorch}, respectively. The training was on a Ubuntu(16.04) desktop with an Intel Core i7 CPU and one GTX 1080Ti NVIDIA GPU and 120 GB memory. We used the DICE loss function and Adam algorithm for optimization, with mini-batch size of 4, weight decay $10^{-4}$. For 2D segmentation, we totally trained 300 epochs for network convergence. The learning rate was kept as $10^{-4}$ for the first 150 epochs and then halved for every 30 epochs. For 3D segmentation, we totally trained 2000 epochs for network convergence. The learning rate was kept as $10^{-4}$ for the first 1000 epochs and then halved for every 200 epochs. Each image/volume was cropped based on the relaxed bounding box derived from the interior margin points firstly and 
then normalized by the mean value and standard deviation of the cropped image. To boost the generalizability to unseen objects, we used several data augmentation methods during the training stage, including random rotation, random scaling, random flipping in space and intensity. Following DeepIGeoS~\citep{Wang18}, we used open source code to compute geodesic distance\footnote{geodesic distance: https://github.com/taigw/GeodisTK.} and solve Eq.~\eqref{eq:eq7} based on max-folw\footnote{max-flow: https://vision.cs.uwaterloo.ca/code/.}, respectively. 
\par In this paper, all testing processes with user interactions were performed on a Ubuntu(16.04) desktop with an Intel Core i7 CPU and a GTX 1080Ti NVIDIA GPU. Following the practice of DeepIGeoS~\citep{Wang18} and BIFSeg~\citep{Wang2018}, the values of $\lambda$ in Eq.~\eqref{eq:eq7} was 5 and $\sigma$ in Eq.~\eqref{eq:eq9} was 0.1 based on a grid search with the validation data. But for specific cases, it also allows the user to set these two parameters manually, like many existing works~\citep{Boykov2002,Rother2004,Criminisi2008}. We developed two PyQt GUIs for user interactions on 2D images and 3D volumes respectively. (See supplementary videos~\footnote{https://www.youtube.com/watch?v=eq-tqlJnckE})

\section{Experiments and results}
\subsection{Comparison methods and evaluation metrics}
To investigate the performance of different encoding methods with the same interior margin points in the first stage of our segmentation method, we compared our EGD with Euclidean distance transform, Gaussian distance transform and geodesic distance transform, which are referred to as EGD, Eucl, Gauss and Geos respectively. In addition, we compared them with segmentation based on the bounding box without encoding of interactions, which is referred to as BBox. All these methods were based on the same CNN structure. For fair comparison, Eucl, Gauss and Geos were implemented with their respective optimal parameters for encoding user-provided interactions. (see the supplementary document)

\par MIDeepSeg was also compared with several existing interactive segmentation methods. In 2D cases, in addition to traditional methods like Graph Cuts~\citep{Boykov2002},  Random Walks~\citep{Grady2006} and SlicSeg~\citep{Wang2016}, we also compared recent deep learning-based methods including  DeepIGeoS~\citep{Wang18}, DIOS~\citep{Xu2016}, DeepGrabCut~\citep{xu2017deep} and DEXTR~\citep{Maninis2017}, where the same 2D network structure was used as our 2D version of MIDeepSeg. For 3D segmentation, we compared MIDeepSeg with ITK-SNAP~\citep{Yushkevich2006} and 3D Graph Cuts~\citep{Boykov2002}, as well as 3D versions of DeepIGeoS~\citep{Wang18}, DIOS~\citep{Xu2016}, DeepGrabCut~\citep{xu2017deep} and DEXTR~\citep{Maninis2017} that used 
the same 3D network as MIDeepSeg for 3D segmentation. Graph Cuts, SlicSeg, Random Walks, DeepIGeoS and DIOS allow the user to refine the results multiple times. DeepGrabCut just allows the user to draw a bounding box at the beginning and does not support further interactions for refinement. DEXTR takes the extreme points as the user interactions and allows the user to refine the results once. Graph Cuts, SlicSeg, Random Walks, and ITK-SNAP are traditional interactive segmentation methods without the need of training with an annotated dataset and have a high generalization. In contrast, DeepIGeoS, DIOS, DeepGrabCut, and DEXTR are deep learning-based methods and require labeled data to train, and DeepIGeoS cannot deal with unseen objects. Two users respectively used these interactive frameworks to segment each test image until the result was visually acceptable, and we reported the average results of the two users achieved. The segmentation results were compared with ground truth label which was annotated by experienced radiologists manually. For quantitative evaluation, we used the Dice similarity coefficient and the average symmetric surface distance (ASSD).

\begin{equation}\label{eq:dice}
    {Dice} = \frac{ 2 \cdot \mid R_p \cap R_g \mid }{\mid R_p\mid + \mid R_g \mid}
\end{equation} where $R_p$ and $R_g$ denote the region segmented by an algorithm and the ground truth label, respectively.

\begin{equation}
    {ASSD} = \frac{1}{\mid S_p \mid + \mid S_g\mid} \bigg(\sum  \limits_{{i} \in {S_p}} d(i, S_g) + \sum  \limits_{{i} \in {S_g}} d(i, S_p)\bigg)
\end{equation} 
where $S_p$ and $S_g$ represent the set of surface points of the result provided by an algorithm and ground truth label, respectively. $d(i, S_p)$ is the shortest Euclidean distance between the point $i$ and the surface $S_g$. To investigate the efficiency of these methods, we listed the user time and amount of user interaction points of each segmentation task.
\subsection{Interactive segmentation of 2D images}
\begin{table}[htb]
    \footnotesize
    \centering
    \scalebox{0.91}
    {\begin{tabular}{lllll}
    \hline
    Object & Modality & Training & Testing & DataSet\\
    \hline
    Placenta & MRI (T2) & 532 slices & 176 slices & Ours\\
    Spleen & CT & 235 slices & 159 slices & BTCV, TCIA$^3$\\
    Kidney & MRI (T1) & No & 100 slices & CHAOS\\
    Kidney & MRI (T2) & No & 100 slices & CHAOS\\
    Kidney & CT & No & 100 slices & CHAOS\\
    Spleen & MRI (T1) & No & 100 slices & CHAOS\\
    Spleen & MRI (T2) & No & 100 slices & CHAOS\\
    Spleen & CT & No & 100 slices & CHAOS\\
    Prostate & MRI (T2) & No & 72 slices & MSD\\
    Fetal brain & Ultrasound & No & 60 slices & HC18\\
    \hline
    \end{tabular}}
    \caption{Datasets used for training and testing the 2D interactive segmentation framework. Note that for spleen, BTCV and TCIA  are training set and testing set, respectively.}
    \label{tab:tab9}
\end{table}
\subsubsection{Data}
Firstly, we validate the propose pipeline with two 2D applications: segmentation of placenta and spleen from fetal MRI and abdomen CT, respectively. Specifically, the placenta data were collected from clinical MRI scans of 30 pregnancies in the second trimester, and were acquired in axial view with pixel size between 0.7422 mm $\times$ 0.7422 mm and 1.582 mm $\times$ 1.582 mm and slice thickness 3 - 4 mm. Each slice was resampled with a uniform pixel size of 1 mm$\times$1 mm. We used 532 slices from 18 volumes, 111 slices from 4 volumes and 176 slices from 8 volumes for training, validation and testing, respectively. The ground truth was manually delineated by an experienced Radiologist. For the spleen data, we randomly selected 235 slices of spleen from 47 volumes (5 slices per volume) in BTCV~\citep{Marsh2013} data set for training, and selected 159 slices from 53 volumes (3 slices per volume) in TCIA\footnote{https://zenodo.org/record/1169361\#.YETa43UzYUE} data set for testing. Secondly, to validate the generalizability of our method, we apply our model trained only with placenta in MRI to four types of organs from a variety of modalities that were not present in the training set: 1) Kidney in CT, T1-weighted and T2-weighted MRI in the CHAOS\footnote{https://chaos.grand-challenge.org} training set. We randomly selected 100 slices for these three cases respectively. 2) Spleen in CT, T1-weighted and T2-weighted MRI in the CHAOS training set. We also randomly selected 100 slices for these three cases respectively. 3) Prostate in T2-weighted MRI from MSD\footnote{http://medicaldecathlon.com/} Task05 dataset,  where we randomly selected 72 slices from 24 cases. 4) Fetal brain in ultrasound images from HC18\footnote{https://hc18.grand-challenge.org/} dataset, where we randomly selected 60 slices. Information of the training and testing set is listed in Tabel~\ref{tab:tab9}. To deal with different organs at different scales, we resized the cropped sub-region and the cue map to 64 $\times$ 64 as the input of CNN.
\begin{figure*}[hptb]
	\centering
	\includegraphics[width=0.75\linewidth]{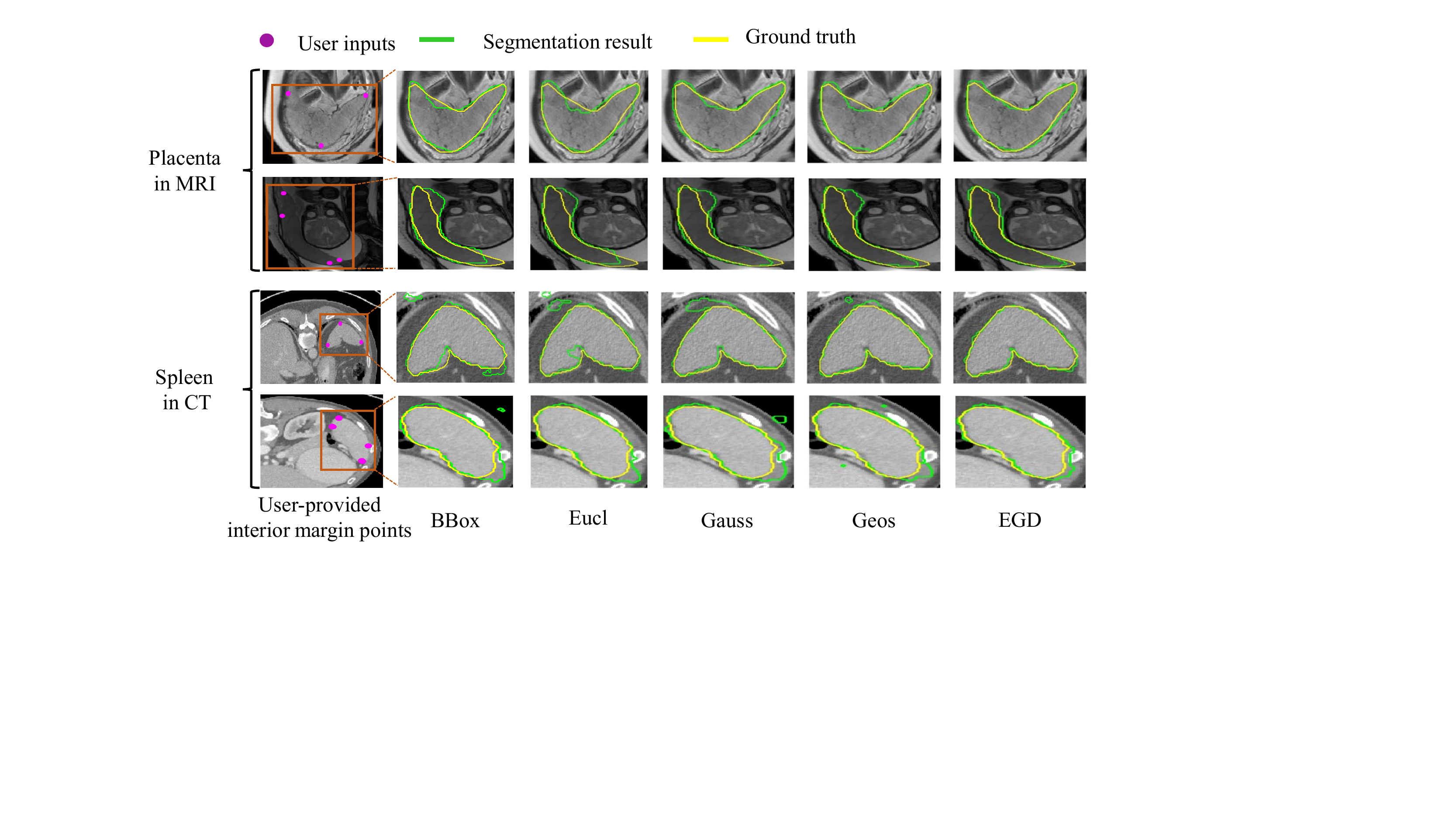}
	\caption{Visual comparison of different encoding methods for placenta and spleen segmentation, in the first stage of our method. The first column shows the input image with user-provided interior margin points (fuchsia). The other column show the initial results.}
	\label{fig:stage-result}
\end{figure*}
\begin{figure*}[hptb]
	\centering
	\includegraphics[width=1.0\linewidth]{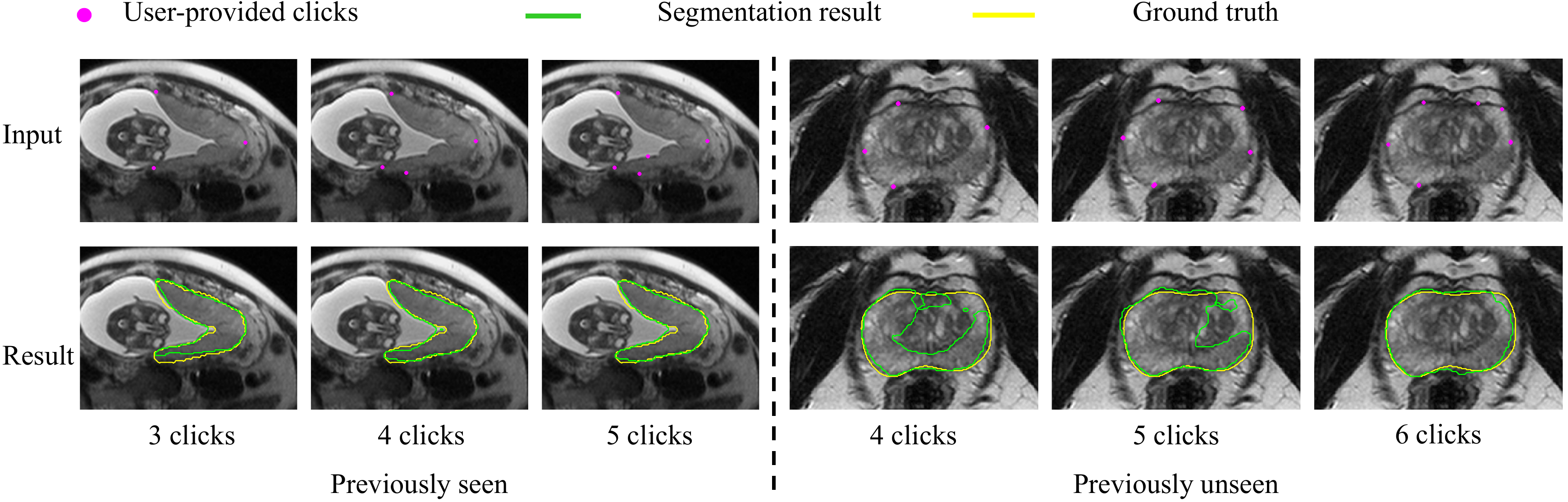}
	\caption{Effect of different number of start interior margin points for segmentation of a placenta (seen object) and prostate (unseen object) with a complex shape. The first row shows the input image with different numbers of interior margin points. The second row shows the segmentation results.}
	\label{fig:different points}
\end{figure*}

\begin{table*}
	\footnotesize
	\centering
	\scalebox{1.0}{\begin{tabular}{lccccc}
			\hline
			{Method}&\multicolumn{2}{c}{ ``Placenta'' from MRI }&\multicolumn{2}{c}{ ``Spleen'' from CT } & \multicolumn{1}{c}{{Time (s)}}\\
			\cline{2-5}
			\multicolumn{1}{c}{} &  Dice ({\%}) & ASS D (pixels) &  Dice ({\%}) & ASSD (pixels) & \\
			\hline
			Bbox &  85.53$\pm$7.16 & 4.37$\pm$3.05 & 91.36$\pm$4.69 & 3.76$\pm$1.71 & -\\
			Eucl &   87.56$\pm$5.98 & 3.42$\pm$2.30 &  93.58$\pm$6.98 & 2.29$\pm$1.13 & \textbf{0.001}\\
			Gauss & 87.91$\pm$6.18 & 3.56$\pm$2.43 & 93.22$\pm$3.32 & 2.29$\pm$1.52 & 0.001\\
			Geos & 87.17$\pm$6.38 & 3.62$\pm$ 1.01 & 94.02$\pm$7.23 &  \textbf{2.13$\pm$0.94} & 0.003\\
			EGD & \textbf{88.10$\pm$4.47}$^*$& \textbf{3.33$\pm$2.19}& \textbf{95.08$\pm$3.23}$^*$ &2.25$\pm$1.28 & 0.004\\\hline
	\end{tabular}}
	\caption{Quantitative comparison of different encoding methods for placenta and spleen segmentation with the same set of interior margin points. The results are based on the initial segmentation of our framework. $^*$ denotes $p$-value $<$ 0.05 when comparing with the second place method.}
	\label{tab:tabn}
\end{table*}
\subsubsection{Initial segmentation based on EGD-based cue map and 2D CNN}\label{subsub:supple}
Fig.~\ref{fig:stage-result} shows some examples of the initial segmentation of placenta from MRI and spleen from CT with user-provided interior margin points, respectively. We compared the proposed EGD with BBox, Eucl (with a threshold), Gauss (with a sigma), Geos (with a threshold) with same user-provided interior margin points, respectively. Note that, the parameters of Eucl, Gauss and Geos were respectively optimized for comparison, and more details of these optimal parameters are listed in Section~\ref{sec:appendix} (appendix). It can be observed that the EGD transform can guide CNN to obtain more accurate segmentation results than the other encoding methods. Table~\ref{tab:tabn} lists the  quantitative  evaluation results of different encoding methods for placenta and spleen. It can be observed that our context-aware and parameter-free encoding method of EGD consistently outperforms the others. The computation time for EGD in 2D is less than 0.05s, which gives real-time response. Fig.~\ref{fig:different points} shows the effect of different number of interior margin points for initial segmentation of challenging cases with complex shapes. 

\begin{figure}
	\centering
	\includegraphics[width=\linewidth]{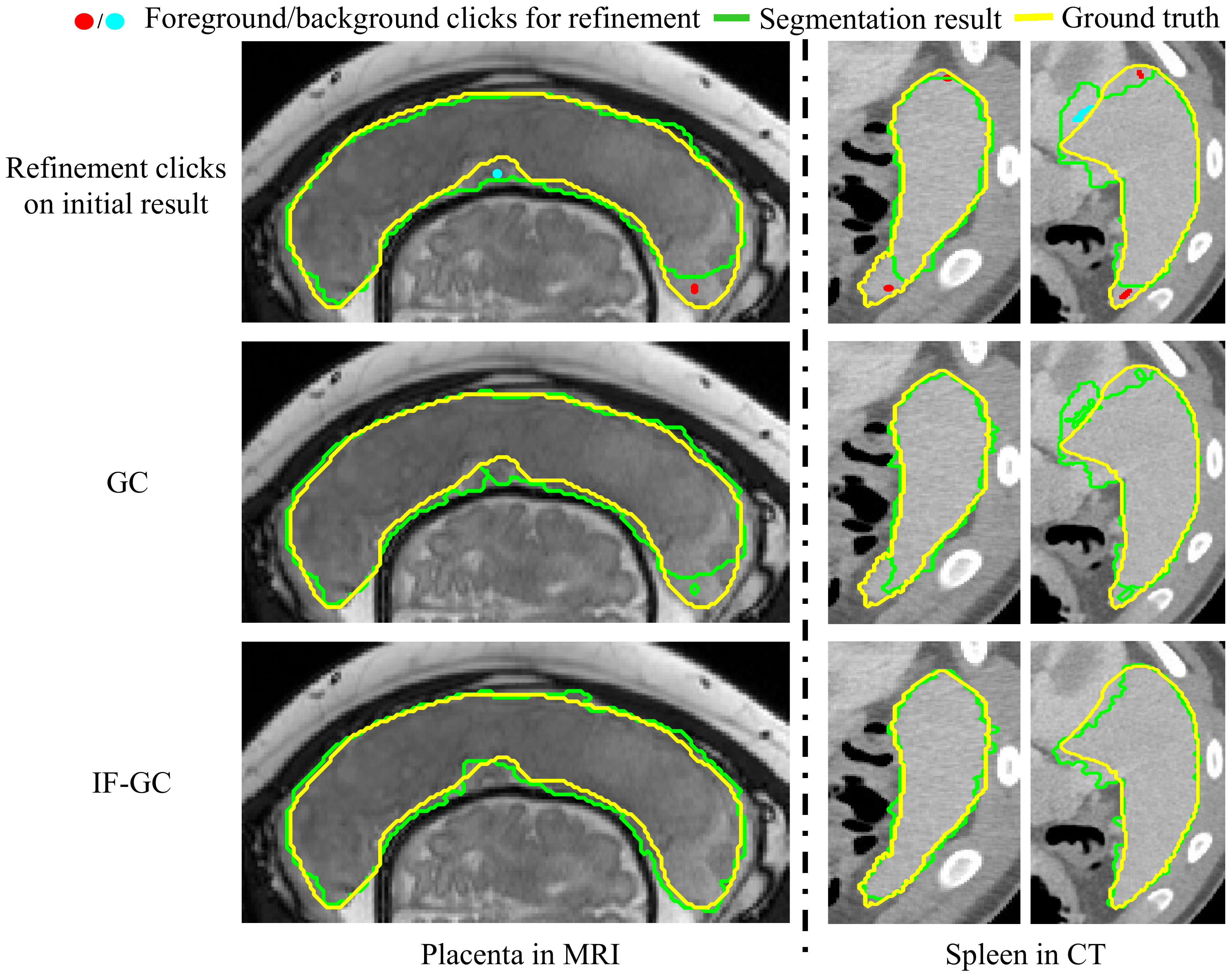}
	\caption{Visual comparison of GC and IF-GC. The first row shows the user clicks for refining initial segmentation result. The other rows show the refined results by GC and IF-GC, respectively. The results are based on the same set of user clicks for refinement. (GC: naive Graph Cuts, IF-GC: information fusion followed by Graph Cut)}
	\label{fig:stage2-result}
\end{figure}
\begin{figure}
	\centering
	\includegraphics[width=1.0\linewidth]{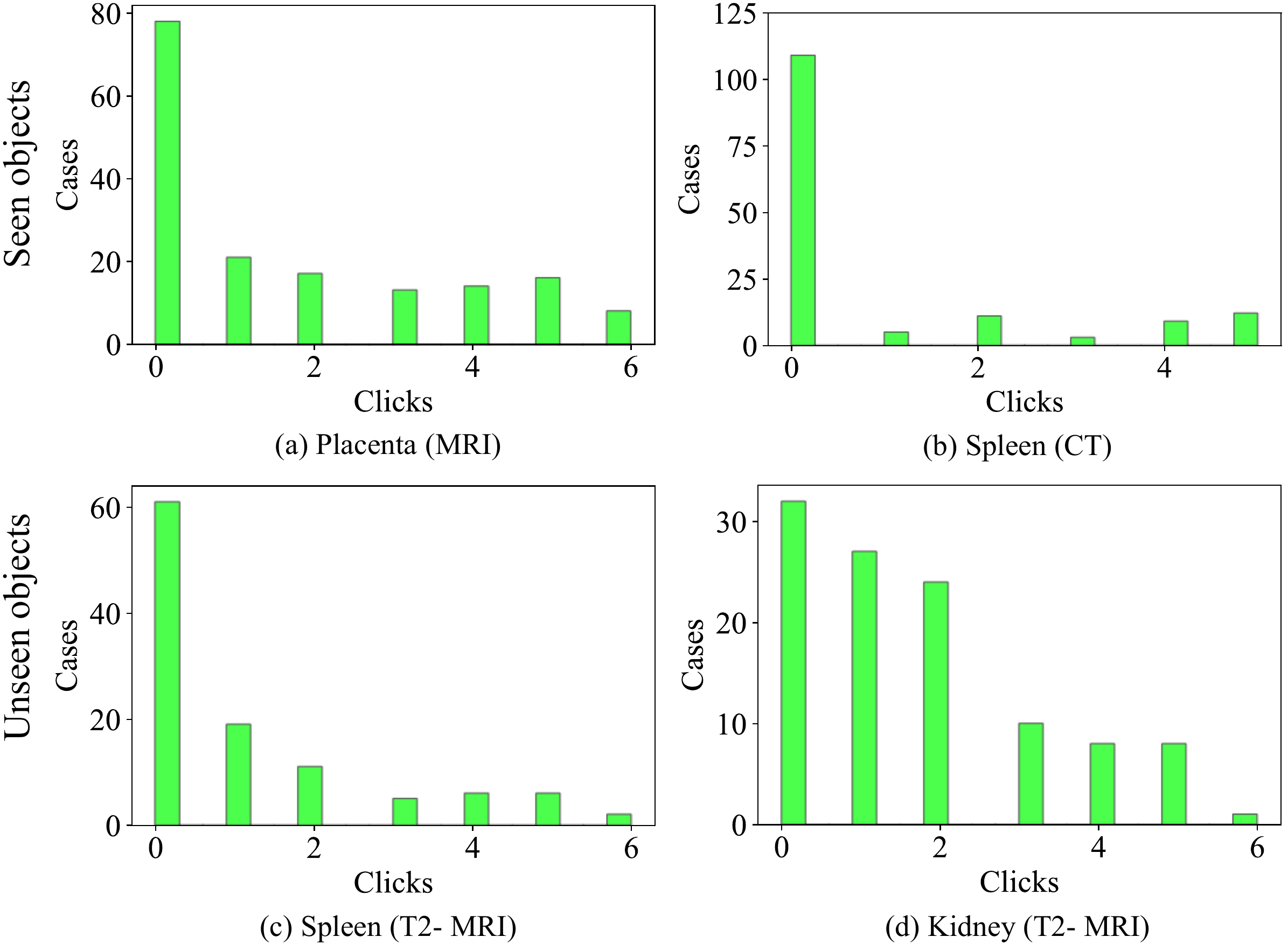}
	\caption{Histogram of number of refinement clicks required by MIDeepSeg for different objects. Placenta (MRI) and spleen (CT) in the first row are seen objects, while spleen (T2-MRI) and Kidney (T2-MRI) in the second row are previously unseen objects.}
	\label{fig:interactions_counts}
\end{figure}
\subsubsection{Interactive refinement by 2D information fusion between initial segmentation and additional clicks}
Fig.~\ref{fig:stage2-result} shows examples of interactive refinement of placenta and spleen segmentation using different refinement methods. The first row shows the initial segmentation obtained in stage 1 of our framework. Based on the initial segmentation, we further use additional clicks to obtain refined results. We compared the refined results between naive Graph Cuts (GC) and information fusion followed by Graph Cuts (IF-GC) using the same set of user clicks. Following the implementation in BIFSeg~\citep{Wang2018}, the naive Graph Cuts takes the initial segmentation probability map and the user interactions (background and foreground seeds) as inputs and is solved by max-flow. A python implementation is publicly available in the SimpleCRF toolkit\footnote{https://github.com/HiLab-git/SimpleCRF}. The performance on placenta and spleen segmentation is listed in Table~\ref{tab:tab2}, where the first two rows show that our method in the first stage already largely outperformed automatic segmentation with the same network structure. The last two rows demonstrate that our IF-GC achieved higher accuracy than naive Graph Cuts with the same set of user clicks for refinement in the second stage.

\par We further investigated the number of refinement clicks for segmentation of different objects using MIDeepSeg and plotted the histogram of refinement click number in Fig.~\ref{fig:interactions_counts}. We can find that a large number of testing cases do not need additional clicks to achieve accurate results and just a few challenging cases need more than 4 clicks for refinement.

\begin{table*}[htb]
    \footnotesize
    \centering
    \scalebox{1.0}{\begin{tabular}{lllll}
    \hline
    \multicolumn{1}{c}{ {Method} }&\multicolumn{2}{c}{ ``Placenta'' from MRI }&\multicolumn{2}{c}{ ``Spleen'' from CT }\\
    \cline{2-5}
    \multicolumn{1}{c}{} &  Dice ({\%}) & ASSD (pixels) &  Dice ({\%}) & ASSD (pixels)\\
    \hline
    Auto & 79.76$\pm$15.33 & 8.94$\pm$11.2 & 90.09$\pm$10.2&10.50$\pm$13.9 \\
    Stage 1 result &88.10$\pm$4.47& 3.33$\pm$2.19& 95.08$\pm$3.23 &2.25$\pm$1.28\\
    Refined by GC &88.41$\pm$5.33 &3.14$\pm$2.39&95.46$\pm$3.19 &2.06$\pm$1.16\\
    Refined by IF-GC& \textbf{89.21}$\pm$\textbf{4.37}$^*$ & \textbf{2.87}$\pm$\textbf{1.89}$^*$ & \textbf{95.79}$\pm$\textbf{3.07}$^*$ & \textbf{1.84}$\pm$\textbf{0.86}\\
    \hline
    \end{tabular}}
    \caption{Quantitative comparison of different refinement methods for placenta and spleen segmentation with the same set of clicks. GC: naive Graph Cuts; IF-GC: information fusion followed by Graph Cuts. * denotes significant difference from GC ($p$-value $<$ 0.05).}
    \label{tab:tab2}
\end{table*}

\begin{table*}[htb]
    \footnotesize
    \centering
    \scalebox{1.0}{\begin{tabular}{lllllllll}
    \hline
    \multicolumn{1}{c}{ {Method} }&\multicolumn{4}{c}{ ``Placenta'' from MRI }&\multicolumn{4}{c}{ ``Spleen'' from CT }\\
    \cline{2-9}
    \multicolumn{1}{c}{} &  Dice ({\%}) & ASSD (pix) & Times (s) & Points (pix) &  Dice ({\%}) & ASSD (pix) & Times (s) & Points (pix) \\
    \hline
    Graph Cuts & 87.02$\pm$5.20 &3.12 $\pm$0.42 &30.1$\pm$10.9 &265.0$\pm$103.6 & 95.27$\pm$4.36&1.30 $\pm$0.42 &21.2$\pm$7.7 &335.1$\pm$91.7 \\
    Random Walks & 87.02$\pm$4.58 &2.95 $\pm$2.66 &33.9$\pm$34.6 &374.3$\pm$114.2 &95.51 $\pm$1.59& 1.45$\pm$2.66 &20.1$\pm$7.9 &218.4$\pm$69.0 \\
    SlicSeg & 87.63$\pm$5.71 &3.00 $\pm$0.39 &25.8$\pm$11.5 &189.3$\pm$81.2 & 95.18$\pm$4.70& 1.23$\pm$0.39 &20.1$\pm$8.2 & 254$\pm$77.5 \\
    DeepIGeoS& 87.96$\pm$5.16 &3.89 $\pm$2.74 &12.0$\pm$8.0 & 90.6 $\pm$ 95.2 &96.39$\pm$2.22&1.71$\pm$2.74 &6.1$\pm$4.8 &31.1$\pm$52.4 \\
    DeepGrabCut& 86.74$\pm$7.03 & 4.18$\pm$2.89 & \textbf{4.2$\pm$2.8}&\textbf{2.0$\pm$0} & 92.54$\pm$3.36 & 2.43$\pm$1.56 & \textbf{3.8$\pm$1.5} & \textbf{2.0$\pm$0} \\
    DIOS& $ 87.48\pm$6.31 & 4.03$\pm$2.52 & 15.3$\pm$13.0& 12.4$\pm$5.7 & 94.85$\pm$2.79 & 2.06$\pm$1.47 & 7.6$\pm$2.7&7.8$\pm$4.7 \\
    DEXTR& 88.77$\pm$4.83 &3.07 $\pm$2.25 & 8.9$\pm$3.7& 7.2$\pm$3.3 & 94.18$\pm$3.25 & 2.67$\pm$1.36 & 5.9$\pm$3.9&5.6$\pm$2.7\\
    MIDeepSeg& \textbf{89.63}$\pm$\textbf{4.15} $^*$ & \textbf{2.69}$\pm$\textbf{1.75} $^*$ &6.40$\pm$3.1 &5.75$\pm$2.1 & \textbf{96.93$\pm$1.43} $^*$ & \textbf{1.18$\pm$0.44} $^*$ & 4.76$\pm$2.0 &4.85$\pm$1.6\\
    \hline
    \end{tabular}}
    \caption{Quantitative comparison of 2D placenta and spleen segmentation by different interactive methods in terms of Dice, ASSD, user time and number of interaction points.  $^*$ denotes $p$-value $<$ 0.05 when comparing with the second place method.}
    \label{tab:seen_diff_tab}
\end{table*}

\subsubsection{Comparison with other interactive methods}
\begin{figure*}[htb]
	\centering
	\includegraphics[width=1.0\linewidth]{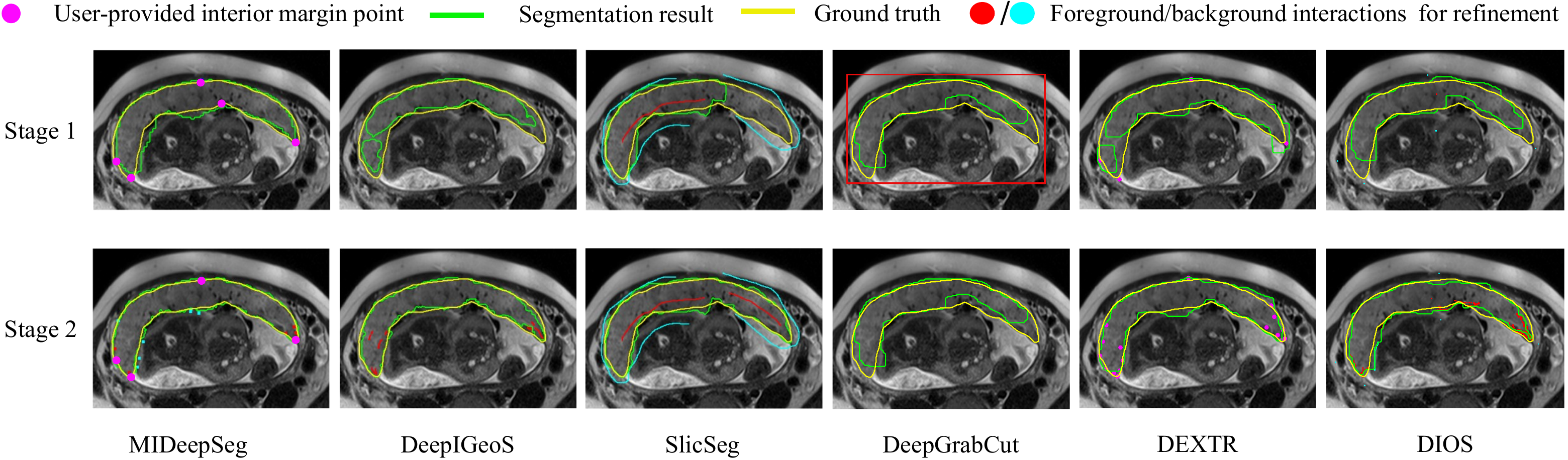}
	\caption{Visual comparison of MIDeepSeg and other interactive methods for 2D placenta segmentation. The first row shows the initial segmentation results with or without initial interactions. And the second row shows the final results after refinement.}
	\label{fig:different-mothds-comparison}
\end{figure*}

We compared MIDeepSeg with DeepIGeoS~\citep{Wang18}, Graph Cuts~\citep{Boykov2002}, Random Walks~\citep{Grady2006}, SlicSeg~\citep{Wang2016}, DIOS~\citep{Xu2016}, DeepGrabCut~\citep{xu2017deep} and DEXTR~\citep{Maninis2017} for placenta and spleen segmentation, respectively. Fig.~\ref{fig:different-mothds-comparison} shows a visual comparison between these methods for 2D placenta segmentation. The first row shows the initial interactions and the initial segmentation results, and the second row shows the final results and all user interactions after refinement. It shows that MIDeepSeg can get a good result with only fewer user clicks, while the others need more interactions. The quantitative comparison of these methods based on placenta and spleen results as presented in Table~\ref{tab:seen_diff_tab}. It shows that MIDeepSeg achieves higher accuracy than the other interactive segmentation methods and it needs less user time and a smaller number of interaction points than the others except for DeepGrabCut. Note that DeepGrabCut does not allow additional user interactions for refinement, which caused the lowest accuracy among the compared methods. This demonstrates that our method is very efficient to obtain highly accurate segmentation results.
 (Also see supplementary video part A)


\subsubsection{Deal with previously unseen 2D objects}
\begin{figure*}[htb]
	\centering
	\includegraphics[width=0.9\linewidth]{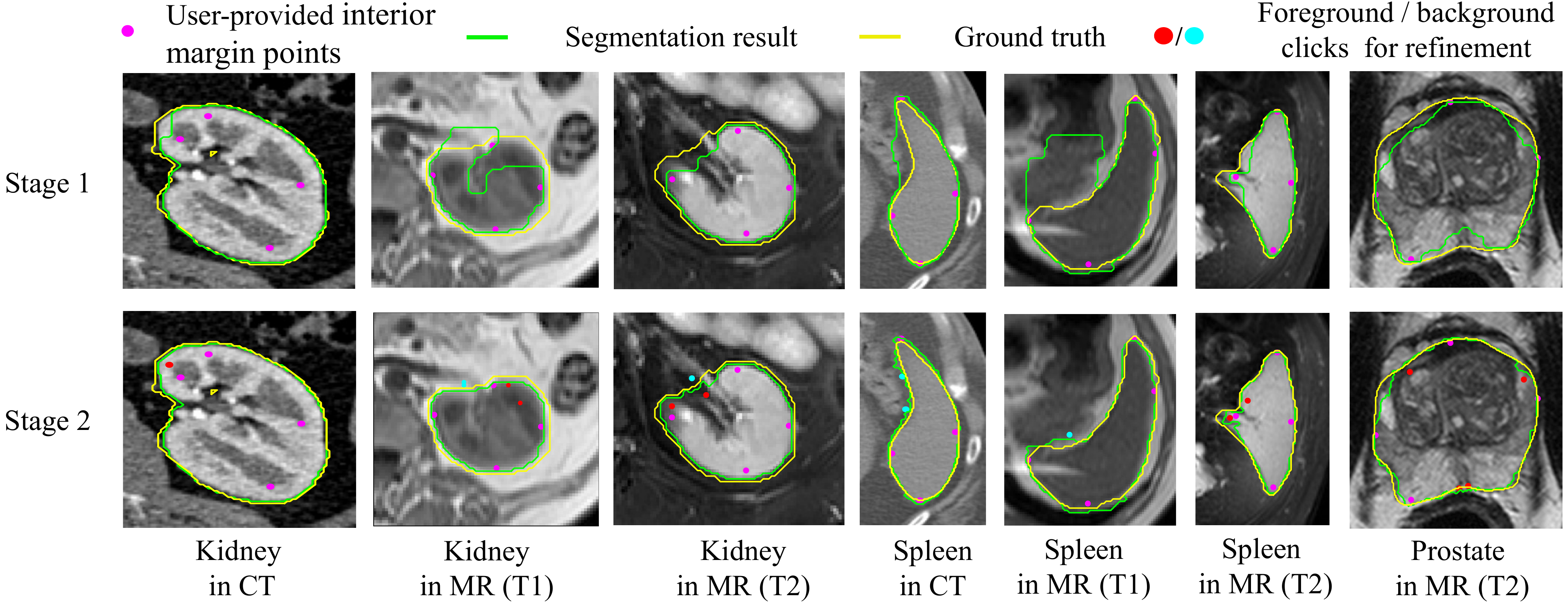}
	\caption{Some examples of 2D unseen organ segmentation results by MIDeepSeg. The first row shows the initial user interactions and the initial segmentation. The second row shows all user interactions and final segmentation results. Note that the model is only trained with placenta in T{2} MRI.}
	\label{fig:2d-unseen-fig}
\end{figure*}

\begin{figure*}
	\centering
	\includegraphics[width=0.7\linewidth]{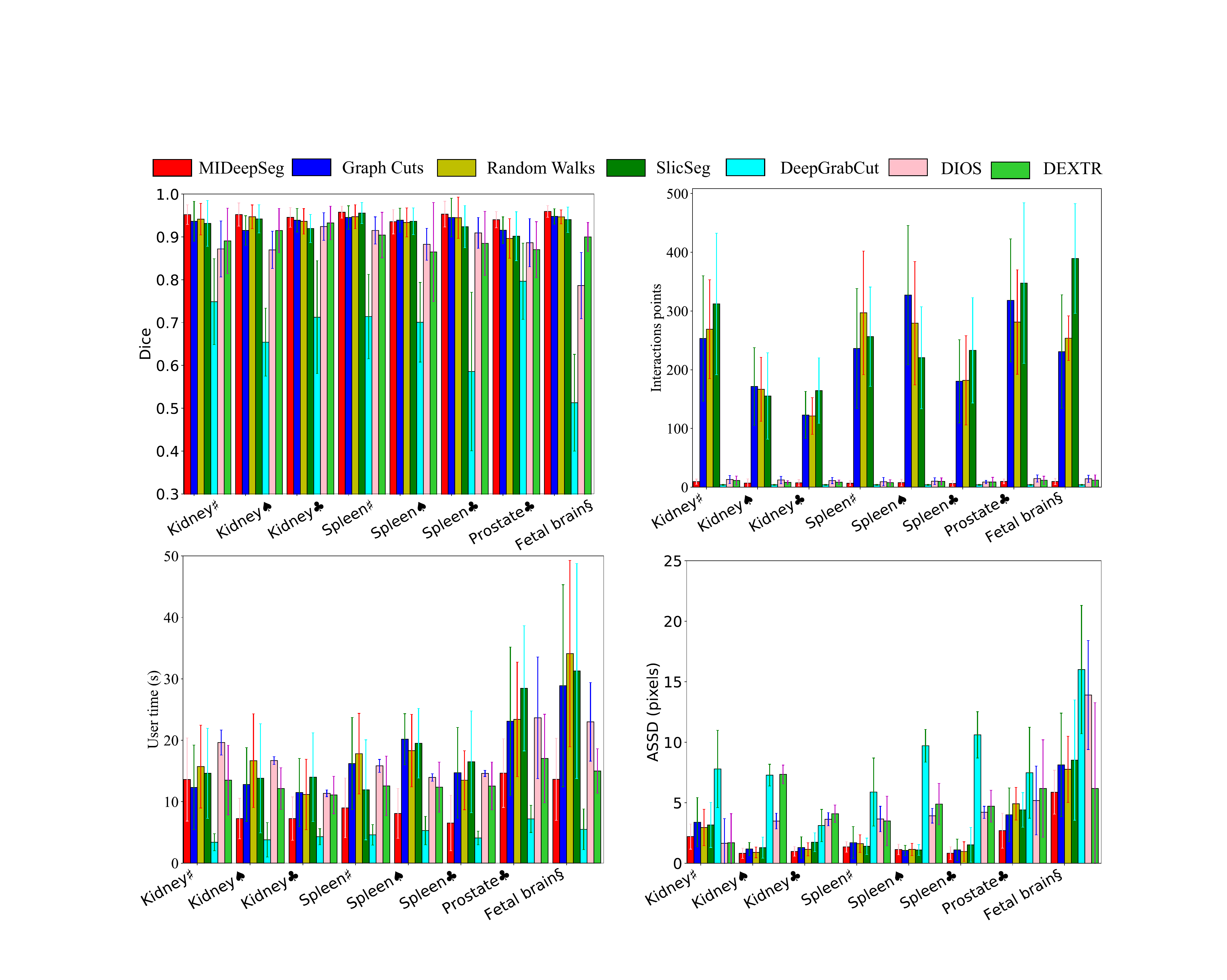}
	\caption{Dice, ASSD, user time and interaction points of different interactive segmentation methods for unseen objects. $\sharp$, $\S$, $\spadesuit$, $\clubsuit$ denote CT images, ultrasound images, T1-weighted MR images and T2-weighted MR images, respectively. All these organs are previously unseen in the training set.}
	\label{fig:2d-unseen-boxplot}
\end{figure*}

To investigate the performance and generalizability of MIDeepSeg on previously unseen objects, we compared MIDeepSeg with existing methods with good generalizability to different objects: Graph Cuts~\citep{Boykov2002}, Random Walks~\citep{Grady2006}, SlicSeg~\citep{Wang2016}, DIOS~\citep{Xu2016}, DeepGrabCut~\citep{xu2017deep} and DEXTR~\citep{Maninis2017}. For MIDeepSeg, DIOS, DeepGrabCut and DEXTR, we used the model that was only trained with placenta images (T2-weighted MRI) to segment four previously unseen organs (i.e., kidney, spleen, prostate and fetal brain) in a range of modalities, as listed in Table~\ref{tab:tab9}. Fig.~\ref{fig:2d-unseen-fig} shows examples of segmentation results of previously unseen objects by MIDeepSeg. The first row shows the initial interactions and initial segmentation results. In the second row, all interactions and final segmentation results are presented. It can be observed that MIDeepSeg can obtain a good result on unseen organs with only few user clicks. The quantitative comparison of these methods based on final results as presented in Fig.~\ref{fig:2d-unseen-boxplot}. It shows that MIDeepSeg takes noticeably less user time and interactions with similar or higher accuracy compared with the other interactive segmentation methods. What is more, it can be observed that MIDeepSeg can deal with different types of previously unseen image modalities and organs very well without any additional training or fine-tuning. We further studied the number of refinement clicks for Kidney (T2-MRI) and spleen (T2-MRI) segmentation using MIDeepSeg and plotted the histogram of refinement click number in Fig.~\ref{fig:interactions_counts}. We can find that although these objects are not present in the training set, our method requires no or only few clicks for refinement to obtain accurate results. (See supplementary video part B)

\subsection{Interactive segmentation of 3D volumes}
\subsubsection{Data}Firstly, we validated the performance of MIDeepSeg on 3D brain tumor core segmentation from contrast-enhanced T1-weighted images. We used the BraTS2018\footnote{https://www.med.upenn.edu/sbia/brats2018.html.} training set which consists of 285 cases with four modalities: FLAIR, T1ce, T1 and T2. All images had been skull-stripped and resampled to an isotropic resolution of 1mm $\times$ 1mm $\times$ 1mm. We used 170 and 47 T1ce cases for training and testing, respectively. Manual segmentations were used as the ground truth.
\par Then, we validated the generalizability of MIDeepSeg with three tasks of segmentation of unseen objects: 1) Whole brain tumor in FLAIR from BraTS2018, from which we randomly selected 60 cases for testing. 2) Kidney in CT from KiTS2019\footnote{https://kits19.grand-challenge.org} dataset, where we randomly selected 15 cases (include 30 kidneys with or without tumor) to test. 3) Left ventricular in MRI from ACDC\footnote{https://acdc.creatis.insa-lyon.fr/}, where we randomly selected 30 cases. The testing data of KiTS and ACDC were resampled to an isotropic resolution of 1mm $\times$ 1mm $\times$ 1mm. All data sets for training and testing are listed in Tabel~\ref{tab:tab3d}. To deal with 3D objects at different scales, we resized the cropped sub-region and cue map to 64 $\times$ 96 $\times$ 96.
\begin{table}[htb]
    \footnotesize
    \centering
    {\begin{tabular}{lllll}
    \hline
    Object & Modality & $N_{train}$ & $N_{test}$ & Dataset\\
    \hline
    Tumor core & MRI (T1ce) & 170  & 47  & BraTS2018\\
    Whole tumor & MRI (FLAIR) & 0 & 60  & BraTS2018\\
    Kidney & CT & 0 & 30  & KiTS\\
    Left ventricular & MRI (T2) & 0 & 30  & ACDC\\
    \hline
    \end{tabular}}
    \caption{Datasets used for training and testing for 3D experiments. $N_{train}$ and $N_{test}$ denote the number of volumes for training and testing, respectively.}
    \label{tab:tab3d}
\end{table}

\subsubsection{Initial segmentation based on EGD-based cue map and 3D CNN}
\begin{figure}[t]
	\centering
	\includegraphics[width=1.0\linewidth]{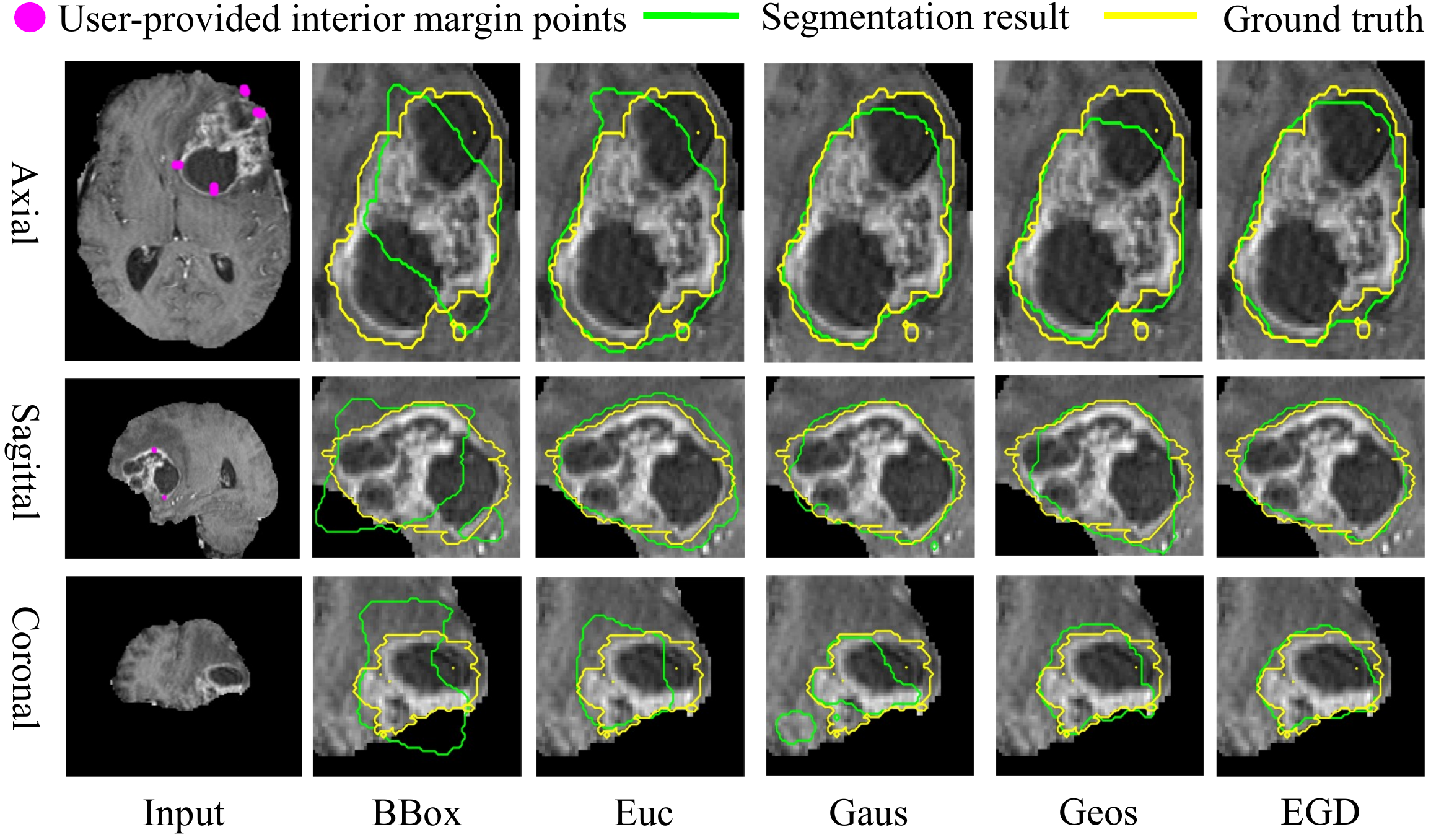}
	\caption{Visual comparison of different encoding methods for 3D tumor core segmentation, which is based on the initial segmentation obtained in the first stage. All these methods used the same interior margin points and inferred bounding box for the input image.}
	\label{fig:3d-encoding-methods}
\end{figure}

To validate our proposed EGD transform for interior margin points encoding in 3D volumes, we compared it with BBox, Eucl, Gauss, and Geos with their respectively optimized parameters, respectively. In this stage, the same set of interior margin points provided by the user were used for these methods. Fig.~\ref{fig:3d-encoding-methods} shows the initial segmentation results obtained by CNN guided by different encoding methods. It shows that EGD transform can guide CNN to achieve more noticeable improvement from BBox compared with the other encoding methods. Table~\ref{tab:tab3} lists the quantitative evaluation results of different encoding methods for tumor core segmentation from T1ce images. It can be observed that our context-aware and parameter-free encoding method of EGD consistently outperforms the others with 87.00\% in term of Dice and 1.46 mm in term of ASSD for tumor core, respectively. Despite that EGD takes more time (0.24s) for interaction encoding than the others, it is still very efficient in practice. 
\begin{table}
	\small
    \centering
    \scalebox{1.0}{\begin{tabular}{lccc}
    \hline
    Method& Dice ({\%}) & ASSD (mm) & Time (s) \\
    \hline
     BBox & 82.32$\pm$12.03 & 2.17$\pm$1.53 & - \\
     Eucl &  85.25$\pm$9.78 & 1.71$\pm$1.20 & \textbf{0.05}\\
     Gauss & 85.90$\pm$9.11 & 1.64$\pm$1.19 & 0.06\\
     Geos & 86.42$\pm$8.91 & 1.60$\pm$1.15 & 0.15 \\
    
     EGD & \textbf{87.00$\pm$9.11}$^*$ & \textbf{1.46$\pm$1.14}$^*$ & 0.24\\
    \hline
    \end{tabular}}
    \caption{Quantitative comparison of different encoding methods for 3D tumor core segmentation with the same set of interior margin points. The results are based on the initial segmentation (Stage 1) of our framework. $^*$ denotes $p$-value $<$ 0.05 when comparing with the second place method.}
    \label{tab:tab3}
\end{table}

\subsubsection{Interactive refinement by 3D information fusion between initial segmentation and additional clicks.}
\begin{figure}[hptb]
	\centering
	\includegraphics[width=1.0\linewidth]{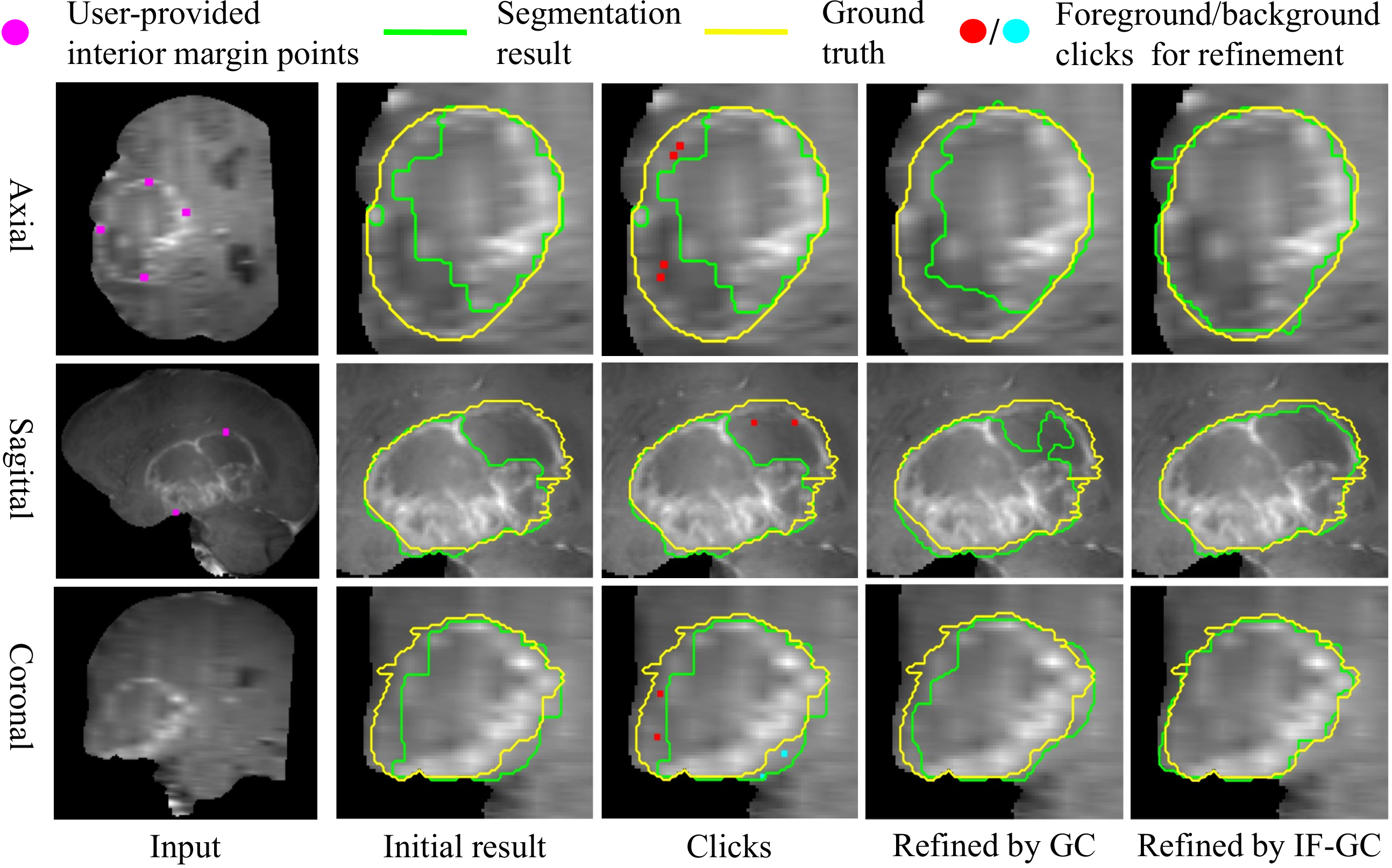}
	\caption{Visual comparison of different refinement methods for 3D tumor core segmentation. These refinement methods are compared with the same initial segmentation with the same set of clicks.}
	\label{fig:3d-refine-methods}
\end{figure}

Based on the above initial segmentation obtained by our method, we further use additional clicks to obtain refined results. We compared the refined result between naive Graph Cuts (GC) and the proposed information fusion followed by Graph Cuts (IF-GC) with the same set of user refinement clicks. The performance on tumor core segmentation is listed in Table~\ref{tab:tab4}, showing that the information fusion achieves higher accuracy than the other variants. Fig.~\ref{fig:3d-refine-methods} shows an example of tumor core segmentation by different refinement methods. It can be observed that IF-GC refined the result more accurately than GC with the same set of clicks for refinement.

\begin{table}[hptb]
    \small
    \centering
    {\begin{tabular}{lll}
    \hline
    Method&  Dice ({\%}) & ASSD (mm)\\
    \hline
    Auto & 78.08$\pm$13.56 & 2.78$\pm$2.22 \\
    Stage 1 result & 87.00$\pm$9.11 & 1.46$\pm$1.14\\
    Refined by GC &87.44$\pm$8.31 &1.37$\pm$1.15\\
    Refined by IF-GC & \textbf{88.21}$\pm$\textbf{7.31}$^*$ & \textbf{1.28}$\pm$\textbf{0.94}\\
    \hline
    \end{tabular}}
    \caption{Quantitative comparison of different refinement methods for 3D tumor core segmentation with the same set of refinement clicks. The segmentation before refinement is obtained by MIDeepSeg in stage 1. GC: 3D Graph Cuts; IF-GC: information fusion followed by Graph Cuts.$^*$ denotes significantly higher performance than GC ($p$-value $<$ 0.05).}
    \label{tab:tab4}
\end{table}

\subsubsection{Comparison with other interactive methods}
\begin{figure}[hptb]
	\centering
	\includegraphics[width=1.0\linewidth]{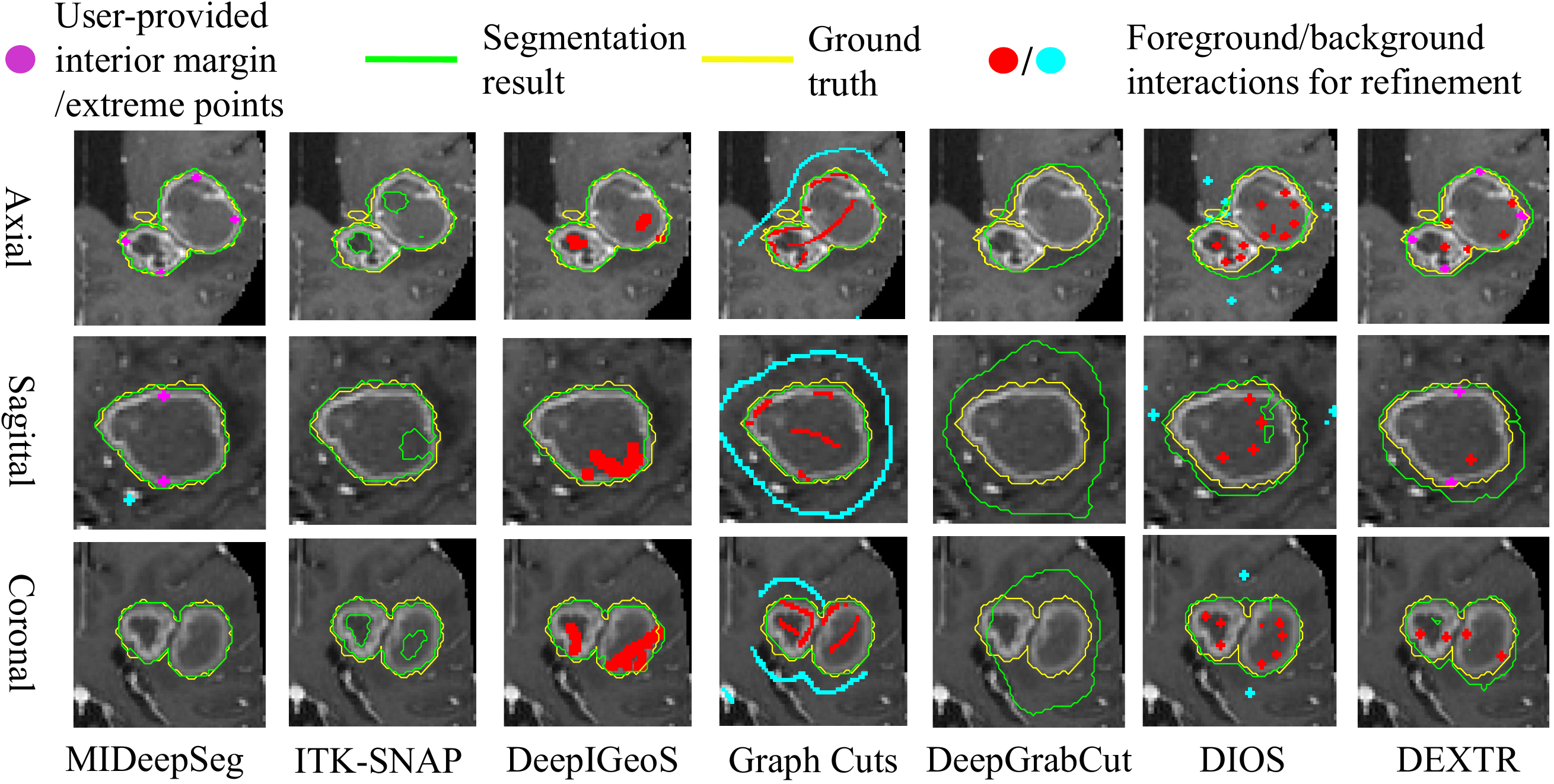}
	\caption{Visual comparison of 3D tumor core segmentation using MIDeepSeg, DeepIGeoS, 3D Graph Cuts and ITK-SNAP.}
	\label{fig:3d-inter-seg-methods}
\end{figure}

Fig.~\ref{fig:3d-inter-seg-methods} shows a visual comparison between MIDeepSeg,  3D Graph Cuts~\citep{Boykov2002}, ITK-SNAP~\citep{Yushkevich2006}, and 3D versions of DeepIGeoS~\citep{Wang18}, DIOS~\citep{Xu2016}, DeepGrabCut~\citep{xu2017deep} and DEXTR~\citep{Maninis2017}. It can be found that MIDeepSeg needs only few interior margin points as the initial interactions, but its initial segmentation is more accurate and it requires fewer user clicks to get an accurate final result. The quantitative comparison of these methods based on the final result is presented in Table~\ref{tab:tab6}. It shows that MIDeepSeg achieved significantly higher accuracy than the others. Additionally, MIDeepSeg takes 29s in average for the entire 3D interactive segmentation process for tumor core, which is less than the other methods except for DeepGrabCut. (See supplementary video part C)


\subsubsection{Deal with previously unseen 3D objects }
\begin{figure*}[hptb]
	\centering
	\includegraphics[width=1.0\linewidth]{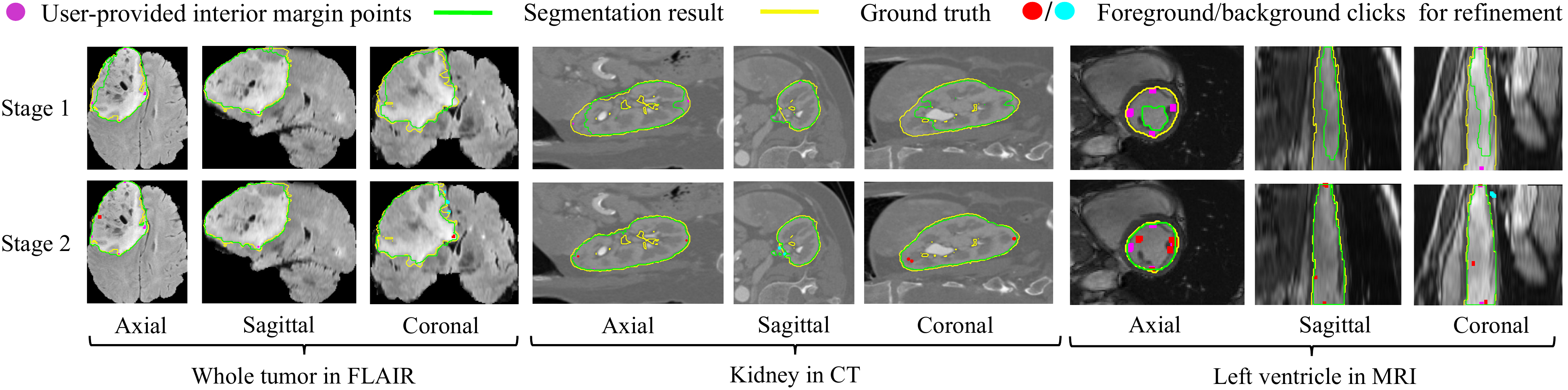}
	\caption{Three examples of segmentation of 3D unseen objects using MIDeepSeg. Note that only tumor core in T1ce images were used for training.}
	\label{fig:3d-unseen-example}
\end{figure*}

To investigate the generalizability of MIDeepSeg on previously unseen 3D objects, we used the 3D CNN model trained in the task of tumor core segmentation from T1ce images to deal with three previous unseen objects and modalities: whole tumor in FLAIR; kidney in CT and left ventricular in MRI, as listed in Table~\ref{tab:tab3d}. Two users used MIDeepSeg and two existing methods with good generalizability including ITK-SNAP~\citep{Yushkevich2006} and 3D versions of Graph Cuts~\citep{Boykov2002},  DIOS~\citep{Xu2016}, DeepGrabCut~\citep{xu2017deep} and DEXTR~\citep{Maninis2017} to segment these objects. Fig.~\ref{fig:3d-unseen-example} shows some examples of 3D whole tumor, kidney and left ventricular segmentation using MIDeepSeg. It can be found that accurate results are obtained for different types of unseen objects by using MIDeepSeg with few clicks. Quantitative evaluation results are presented in Fig.~\ref{fig:comparison-unseen-box}. It shows that MIDeepSeg achieves similar or higher accuracy compared with 3D Graph Cuts, ITK-SNAP, DeepIGeoS, DIOS, DeepGrabCut and DEXTR. However, MIDeepSeg takes notable less user time to achieve the results. (See supplementary video part D)

\begin{table}[hptb]
	\scriptsize
    \centering
    {\begin{tabular}{llll}
    \hline
    Method&  Dice ({\%}) & ASSD (mm) & Time (s)\\
    \hline
    3D Graph Cuts & 78.91$\pm$14.98 & 3.46$\pm$5.10 & 99.4$\pm$36.7 \\
    ITK-SNAP & 82.34$\pm$11.42  &1.99 $\pm$1.31 & 173.0$\pm$75.5 \\
    DeepIGeoS & 82.47$\pm$10.78  & 3.34$\pm$3.81  & 82.2$\pm$44.7 \\
    DeepGrabCut & 82.58$\pm$11.79 & 2.89$\pm$2.37 & \textbf{10.5$\pm$8.3} \\
    DIOS & 83.34$\pm$10.47  & 2.57$\pm$1.79 & 67.5$\pm$23.6 \\
    DEXTR & 86.39$\pm$9.03  & 1.59$\pm$1.11 & 34.7$\pm$18.6\\
    MIDeepSeg & \textbf{88.71$\pm$7.00}$^*$ &\textbf{1.24 $\pm$0.88}$^*$ &28.6 $\pm$12.2 \\
    \hline
    \end{tabular}}
    \caption{Quantitative evaluation of 3D tumor core segmentation by different interactive methods in terms of Dice, ASSD and user time, respectively. $^*$ denotes $p$-value $<$ 0.05 when comparing with the second place method.}
    \label{tab:tab6}
\end{table}

\section{Discussion}
Though some recent works~\citep{Wang18,zhou2019interactive, wang2020uncertainty, Liao_2020_CVPR} on deep learning-based interactive segmentation have shown good performance, it is a great challenge for current CNNs to generalize well on previously unseen object classes, as they rely on annotated image to learn directly~\citep{masood2015survey}. For medical images, annotated images are very precious and scarce since accurate annotations require both expertise and time to obtain. This limits the performance of CNNs to deal with unseen objects that are not present in training set. Compared with traditional CNNs~\citep{Ronneberger2015, Cicek2016} and transfer learning~\citep{Wang2018, Tajbakhsh2016}, the major advantage of ours proposed framework is that it can segment unseen objects without re-training or fine-tuning. Therefore, it reduces the burden for collecting and annotating data noticeably, and can be applied to segment or annotate unseen objects directly. Compared with DeepIGeoS~\citep{Wang18} and BIFSeg~\citep{Wang2018}, MIDeepSeg only requires few clicks as input and has a higher generalizability.
\par A big challenge for existing deep learning frameworks is that they hardly generalize on previously unseen objects, and they require additional re-training or fine-tuning for segmentation of new targets. BIFSeg~\citep{Wang2018} uses image-specific fine-tuning to improve the generalization of CNN, but it requires fine-tuning for each test image, which is a time and memory consuming process. Based on our proposed interior margin points, EGD transform and information fusion, MIDeepSeg can deal with different types of unseen medical images without additional fine-tuning or training. 
\begin{figure*}[hptb]
	\centering
	\includegraphics[width=0.8\linewidth]{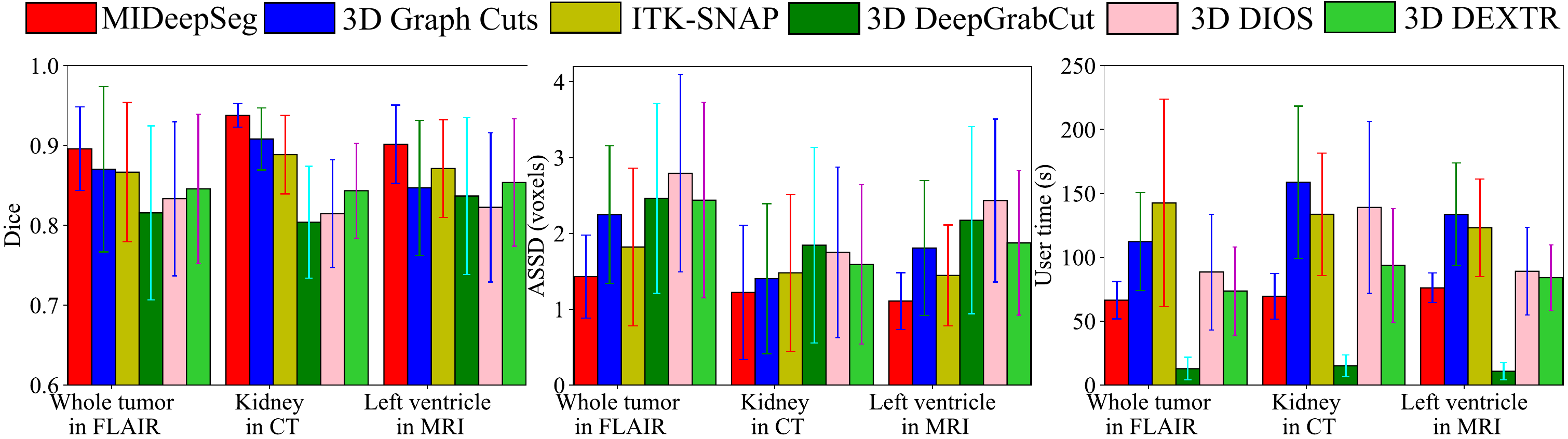}
	\caption{Dice score, ASSD and user time of different interactive methods for 3D unseen objects segmentation. }
	\label{fig:comparison-unseen-box}
\end{figure*}

Despite the simple implementation, our EGD has not been proposed earlier for user interaction encoding, and it has two important differences from geodesic distances: First, EGD is parameter-free with higher generalizability. The geodesic distance method~\citep{Wang18} requires a user-defined threshold to make sure that the interactions will affect a local region, which reduces its generalizability as different images may require different threshold values. In contrast, our EGD does not require such a parameter, and it can be applied to different images without some specific adjustment, making it a simple and effective method with a wider utility. Second, the EGD naturally outputs a probabilisticty map, which gives can be used as the probability of a pixel belonging to the foreground or background indicated by the user interactions. This probabilistic view allow it to be seamlessly integrated in to a conditional random field formulation for refinement.

The computation time of our EGD listed in Table~\ref{tab:tabn} and Table~\ref{tab:tab3} show that it takes less than 0.05s and 0.25s for 2D and 3D images respectively, which is acceptable for fast response of user interactions. We also studied the computation time of other stages of our method: the inference time for 2D and 3D CNNs was 0.008s and 0.04s, respectively. The CRF optimization time was 0.015s and 0.5s for 2D and 3D images, respectively. The entire user time was around 8-12s for 2D cases (Fig.~\ref{fig:2d-unseen-boxplot}) and 60-80s for 3D cases (Fig.~\ref{fig:comparison-unseen-box}). Therefore, our method is efficient for interactive segmentation of unseen objects.

\par In our experiments, we found that our refinement method based on calibrated probability maps and Graph Cuts worked well in various cases for different organs in a range of modalities. The advantages include: 1) the refinement step is decoupled from the initial segmentation step based on CNNs, thus is ready-to-use as a general refinement tool for interactively correcting segmentation results obtained by different networks and for unseen objects. 2) It is computational efficient, and allows real-time response of user interactions, which is highly desired for improving the user experience of interactive segmentation. 3) The user interactions are used as hard constraints, which ensures that points given by users will have their desired labels after refinement. A potential issue is that in complex cases a relatively large number of clicks are needed to obtain accurate results. However, in practice, our method is easy to use and efficient in dealing with different unseen objects, as shown by the experimental results.

A general problem for interactive segmentation is that the result may depend on knowledge and experience of the user, as the user refines the segmentation until it is visually acceptable, where the standard may be subjective.  However, our method has some requirements on the user interactions: in the first stage, the interactions need to be given near the inner side boundary, and in the second stage the interactions are only given in incorrect regions, where for most cases, the incorrect region is small, leading to the range of clicks provided by different users limited. Therefore, inter-user variation of our method is small.  As our method does not require the user to provide clicks exactly on the boundary or extreme points, our interior margin points tolerate inaccurate clicks, which is more user-friendly. As shown in Fig.~\ref{fig:2d-unseen-fig}, in the first column the interior margin points are not accurate and far away from the boundary, and in the fourth column the top point is also inaccurate and even clicked in the background, but they still lead to good initial segmentation results. It further demonstrates the robustness and generalization of MIDeepSeg.

Recently, some works~\citep{Sourati2019, Rupprecht2018, Zhou2019, Lee2018} used Fisher information, natural language, active learning and deep reinforcement learning to develop an intelligent interactive segmentation or annotation tool. In the future, it is of interest to use active learning~\citep{top2011active}and deep reinforcement learning~\citep{Liao_2020_CVPR} and uncertainty estimation~\citep{wang2020uncertainty} to provide guidance on user interactions for refinement, which has a potential to further improve the efficiency of interactive segmentation. 

\section{Conclusion}
In this paper, we proposed a deep learning-based interactive framework with good generalizability to unseen objects for medical images segmentation and it only requires few clicks as user inputs. A novel context-aware and parameter-free encoding method was proposed to encode user interactions to guide CNN for a good initial segmentation. Based on the encoding method, we also proposed an effective refinement way for improving the accuracy of the segmentation results. The framework is designed to improve the generalizability to unseen objects, which is highly desired for deep learning-based models. Experiments on segmenting a wide range of previously seen and unseen organs or lesions from various 2D and 3D images show that: 1) Our interior margin points and EGD transform-based framework outperforms existing deep learning-based interactive segmentation tools in terms of accuracy and efficiency. 2) The proposed framework generalizes well on previously unseen objects. It could be used as an annotation tool to obtain segmentation masks of a range of objects more efficiently with high accuracy.

\section{Acknowledgment}
This work was supported by the National Natural Science Foundations of China [61901084 and 81771921] funding, key research and development project of Sichuan province, China [No. 20ZDYF2817]. This work also was supported by the Wellcome Trust [WT101957, 203148/Z/16/Z], and the Engineering and Physical Sciences Research Council (EPSRC) [NS/A000027/1, NS/A000049/1]. TV is supported by a Medtronic / Royal Academy of Engineering
Research Chair [RCSRF18194].
\section{Supplementary Video}
The supplementary video of this paper can be found at: \url{https://www.youtube.com/watch?v=eq-tqlJnckE}.
\begin{table*}
	\small
	\centering
	{\begin{tabular}{ccccc}
			\hline
			\multicolumn{1}{c}{ {Threshold} }&\multicolumn{2}{c}{ ``Placenta'' from MRI }&\multicolumn{2}{c}{ ``Spleen'' from CT }\\
			\cline{2-5}
			\multicolumn{1}{c}{} &  Dice ({\%}) & ASSD (pixels) &  Dice ({\%}) & ASSD (pixels)\\
			\hline
			
			0.2 &  86.82$\pm$6.65 & 3.74$\pm$2.58 & 92.88$\pm$3.81 & 2.47$\pm$1.58\\
			
			0.4 & 86.91$\pm$7.20 & 3.62$\pm$2.47 &  \textbf{93.58$\pm$6.98} & \textbf{2.29$\pm$1.13}\\
			
			0.6 &   \textbf{87.56$\pm$5.98} & \textbf{3.42$\pm$2.30} &  92.62$\pm$3.13 & 2.30$\pm$1.16\\
			
			0.8 & 87.13$\pm$6.26 & 3.57$\pm$ 2.29 & 93.04$\pm$5.00 &  2.33$\pm$1.08  \\
			\hline
	\end{tabular}}
	\caption{Quantitative evaluation of different threshold values in Euclidean distance transforms for placenta and spleen segmentation with the same set of real-users clicked points. The results are based on the initial segmentation of our framework.}
	\label{tab:euc-different-para}
\end{table*}
\begin{table*}
	\small
	\centering
	\scalebox{1.0}{\begin{tabular}{ccccc}
			\hline
			\multicolumn{1}{c}{ {$\sigma$} }&\multicolumn{2}{c}{ ``Placenta'' from MRI }&\multicolumn{2}{c}{ ``Spleen'' from CT }\\
			\cline{2-5}
			\multicolumn{1}{c}{} &  Dice ({\%}) & ASSD (pixels) &  Dice ({\%}) & ASSD (pixels)\\
			\hline
			
		3 &  86.96$\pm$6.89 & 3.74$\pm$2.58 & 92.76$\pm$4.16 & 2.38$\pm$1.52\\
			
			6 & 87.78$\pm$6.15 & 3.62$\pm$2.47 &  92.91$\pm$6.98 & 2.21$\pm$1.45\\
			
			9 &  \textbf{87.91$\pm$6.18} & \textbf{3.42$\pm$2.30} &  \textbf{93.22$\pm$3.32} & \textbf{2.29$\pm$1.51}\\

			12 & 86.97$\pm$6.79 & 3.57$\pm$ 2.29 & 92.69$\pm$3.70 &  2.24$\pm$1.26  \\
			\hline
		\end{tabular}
	}
	\caption{Quantitative evaluation of different $\sigma$ in Gaussian distance transforms for placenta and spleen segmentation with the same set of real-users clicked points. The results are based on the initial segmentation of our framework.}
	\label{tab:gaus-different-para}
\end{table*}

\begin{table*}
	\small
	\centering
	\scalebox{1.0}{\begin{tabular}{ccccc}
			\hline
			\multicolumn{1}{c}{ {Threshold} }&\multicolumn{2}{c}{ ``Placenta'' from MRI }&\multicolumn{2}{c}{ ``Spleen'' from CT }\\
			\cline{2-5}
			\multicolumn{1}{c}{} &  Dice ({\%}) & ASSD (pixels) &  Dice ({\%}) & ASSD (pixels)\\
			\hline
			
			0.2 &  86.81$\pm$6.33 & 4.00$\pm$2.72 & 92.96$\pm$3.13 & 2.23$\pm$0.97\\
			
			0.4 & 87.08$\pm$5.35 & 3.59$\pm$2.05 &  \textbf{94.02$\pm$3.23} & \textbf{2.13$\pm$0.93}\\
			
			0.6 &   \textbf{87.17$\pm$6.38} & \textbf{3.55$\pm$2.62} &  93.06$\pm$4.67 & 2.17$\pm$0.96\\

			0.8 & 86.93$\pm$7.13 & 3.62$\pm$ 2.45 & 92.91$\pm$3.33 &  2.23$\pm$0.92  \\
			\hline
	\end{tabular}}
	\caption{Quantitative evaluation of different thresholds in Geodesic distance transforms for placenta and spleen segmentation with the same set of real-users clicked points. The results are based on the initial segmentation of our framework.}
	\label{tab:geos-different-para}
\end{table*}

\section{Appendix}\label{sec:appendix}
In Section~\ref{subsub:supple}, we compared our parameter-free interaction encoding method EGD with several existing encoding methods: Euclidean distance transform,  Gaussian heatmap and Geodesic distance transform. Each of these alternative methods relies on a hyper-parameter. Here we report the details of finding the optimal hyper-parameter values for these methods. The dataset and CNN structure were the same as those used in Section 3.2.2.
\subsection{Effects of different threshold for Euclidean distance transforms}
DeepIOS~\citep{Xu2016} and ISLD~\citep{Li2018} use euclidean distance to encode user interactions. The distance transforms are truncated by a threshold value for efficient representation. Then, the truncated encoding map is concatenated with the input image to guide the CNN to obtain a segmentation result. Following these works, we rescaled all encoding maps to $\left [0, 1  \right ]$ and truncated them with  four different threshold values: 0.2, 0.4, 0.6 and 0.8, respectively. We used their corresponding cue maps to guide CNN to segment placenta in MRI and spleen in CT, respectively. Table~\ref{tab:euc-different-para} lists the corresponding quantitative evaluation results. We found that the optimal threshold value was 0.6 for the placenta, and 0.4 for the spleen. We therefore used these optimal values for DeepIOS~\citep{Xu2016} and ISLD~\citep{Li2018} in the experiment.

\subsection{Effects of different sigma for Gaussian distance transforms}
DEXTR~\citep{Maninis2017} and DELSE~\citep{Cvpr2019} use Gaussian distance transforms to deal with user interactions. To generate the gaussian heat-map, we need to determine a proper sigma for each task. In this work, we used four different values of sigma (i.e., 3, 6, 9 and 12) to generate gaussian heatmap for placenta and spleen segmentation, respectively.  Their corresponding heatmaps were concatenated with the input image to guide the CNN to achieve an initial segmentation. The results are listed in Table~\ref{tab:gaus-different-para}. It can be found the optimal sigma value was 9 for the placenta and the spleen. We therefore used these optimal values for DEXTR~\citep{Maninis2017} and DELSE~\citep{Cvpr2019} in the experiment.
\subsection{Effects of different threshold for Geodesic distance transforms}
DeepIGeoS~\citep{Wang18} encodes user interactions by geodesic distance transforms. Similarly to Euclidean distance transforms, we truncated the geodesic distance maps with different threshold values: 0.2, 0.4, 0.6 and 0.8, respectively. These geodesic distance maps were concatenated with the input image to guide the CNN to achieve an initial segmentation. The quantitative evaluation results are listed in Table~\ref{tab:geos-different-para}. It can be observed the optimal threshold value was 0.6 for the placenta, and 0.4 for the spleen. We therefore used these values for DeepIGeoS~\citep{Wang18} in the experiment.
\bibliographystyle{model2-names.bst}\biboptions{authoryear}
\bibliography{refs}
\end{document}